# GCL-GCN: Graphormer and Contrastive Learning Enhanced Attributed Graph Clustering Network

***Binxiong Li[a], Xu Xiang[b], Xue Li[b], Quanzhou Luo[a], Binyu Zhao[a], Yujie Liu[a], Huijie Tang[a], Benhan Yang[a]***

[a]*Stirling College, Chengdu University, 2025 Chengdu Rd., Chengdu, Sichuan, China, 610106*

[b]*College of Computer Science, Chengdu University, 2025 Chengdu Rd., Chengdu, Sichuan, China, 610106*

**ABSTRACT**

Attributed graph clustering holds significant importance in modern data analysis. However, due to the complexity of graph data and the heterogeneity of node attributes, leveraging graph information for clustering remains challenging. To address this, we propose a novel deep graph clustering model, GCL-GCN, specifically designed to address the limitations of existing models in capturing local dependencies and complex structures when dealing with sparse and heterogeneous graph data. GCL-GCN introduces an innovative Graphormer module that combines centrality encoding and spatial relationships, effectively capturing both global and local information between nodes, thereby enhancing the quality of node representations. Additionally, we propose a novel contrastive learning module that significantly enhances the discriminative power of feature representations. In the pre-training phase, this module increases feature distinction through contrastive learning on the original feature matrix, ensuring more identifiable initial representations for subsequent graph convolution and clustering tasks. Extensive experimental results on six datasets demonstrate that GCL-GCN outperforms 14 advanced methods in terms of clustering quality and robustness. Specifically, on the Cora dataset, it improves ACC, NMI, and ARI by 4.94%, 13.01%, and 10.97%, respectively, compared to the primary comparison method MBN.

**The source code for this study is available at https://github.com/YF-W/GCL-GCN**

**Corresponding author**: Binxiong Li
**E-mail**: libinxiong@stu.cdu.edu.cn

## 1. Introduction

In modern data analysis, clustering is a fundamental and significant task aimed at categorizing similar samples into the same group. With the remarkable breakthroughs in deep learning within the field of machine learning, deep clustering methods have become mainstream due to their superior feature extraction capabilities and flexibility, making them indispensable in practical applications. In this context, Yan et al. proposed the Mutual Boost Network (MBN), which combines AE and graph autoencoders (GAE) to extract information from both node and structural features, achieving significant results in graph clustering tasks [1]. Zhang et al. introduced the Transformer-based Dynamic Fusion Clustering Network (TDCN), which incorporates the Transformer architecture with AE networks and employs a novel dynamic attention mechanism to efficiently fuse multiple network features. This network utilizes a dual self-supervised mechanism for training, significantly enhancing clustering performance [2]. In recent years, graph contrastive learning has demonstrated notable improvements in handling graph-structured data by leveraging intrinsic relationships between nodes. For instance, Wang et al. proposed the Dual Contrastive Attributed Graph Clustering Network (DCAGC), which integrates contrastive learning directly with clustering tasks, effectively improving the quality of node representations and optimizing training within a unified framework, thus embedding clustering-oriented information in the learned node representations [3]. Additionally, Joshua et al. presented Contrastive Learning of Hard Negative Samples, which combines contrastive learning with deep graph clustering to form tight clusters in the embedding space and separate different clusters as much as possible. This paper introduces a new unsupervised sampling method to select hard negative samples, thereby enhancing the performance of contrastive learning and, consequently, achieving better clustering [4]. These models, through advanced deep learning and contrastive learning techniques, have shown significant improvements and importance in handling graph-structured data, greatly enhancing clustering performance and application value.

GCN method has demonstrated exceptional performance in graph data clustering tasks. Clustering methods based on GCN are highly regarded for their ability to effectively capture graph structural information. Kipf and Welling (2016) proposed the GAE model, which learns node representations in an unsupervised manner and captures relationships between nodes by reconstructing the graph structure [5]. More recently, Tsitsulin et al. introduced the Graph Cluster Neural Network (GCNN) model, which enhances node clustering by optimizing node embeddings with a clustering loss function [6]. The Transformer model, proposed by Vaswani et al. (2017), has achieved significant success in natural language processing and has gradually been applied to graph data clustering tasks [7]. The Graph Transformer Network (GTN) model (2022) designs specialized graph structure encoders and decoders, leveraging self-attention mechanisms to capture global dependencies, thus demonstrating significant advantages in handling large-scale and complex graph data [8]. Furthermore, contrastive learning, as a self-supervised learning method, has shown tremendous potential in feature learning and clustering tasks in recent years. The Contrastive Clustering (CC) model (2021) constructs multi-level contrastive learning tasks to incrementally optimize node embeddings, effectively handling noisy datasets and significantly improving clustering robustness [9]. You et al. (2022) proposed the Deep Graph Clustering via Dual Correlation Reduction (DCR) model, which achieves effective clustering by maximizing the correlation between node feature representations and structural representations. Contrastive learning methods offer significant advantages in enhancing feature representation quality and clustering performance [10].

**Motivation and Contribution**

With the notable advantages demonstrated by GCN in processing graph data, researchers have increasingly recognized the potential of these methods in capturing graph structural information. However, existing GCN methods still face numerous challenges when dealing with highly sparse and heterogeneous graph data. Specifically, traditional GAE performs poorly in reconstructing graph structures and node features, while GCNN, in focusing too much on node aggregation, overlooks fine-grained information when handling complex graph structures. Although Transformer-based graph models excel at capturing global dependencies, they often struggle with information insufficiency when dealing with sparse graph data, resulting in inadequate local dependency capture. Moreover, while contrastive learning shows certain advantages in feature representation learning, shallow architectures fail to effectively capture deep graph structural information, and complex architectural designs may amplify noise within the data.

To address these challenges, this study proposes a novel deep graph clustering model, GCL-GCN. By leveraging the strengths of the AE module, GCN module, newly encoded Graphormer module, and an innovative contrastive learning module, this model effectively enhances the clustering performance of graph data.

**Contributions of GCL-GCN**

- **Modular Design with a Multi-Task Joint Learning Framework:** We have designed a modular model framework that organically integrates the contrastive learning module, AE module, GCN module, and Graphormer module of our model. By employing a joint training strategy, this approach effectively combines the advantages of various modules, enabling efficient extraction and clustering of graph data information.

- **Innovative Encoding with Centrality Features and Spatial Relationships:** By introducing node centrality features and spatial relationships in the Graphormer module, our model can better capture both local and global information of the nodes, thereby enhancing node representation capability and improving clustering performance.

- **Application of Innovative Contrastive Learning Pretraining Technique:** In the pretraining phase, we employ contrastive learning techniques, inputting the original feature matrix to significantly enhance the discriminative power of feature representations through contrastive learning. This method effectively increases feature distinction during pretraining, providing more identifiable initial feature representations for subsequent graph convolution and clustering tasks.

- **Joint Optimization of Reconstruction Error and Contrastive Learning Loss:** By jointly optimizing the reconstruction error of the AE module, the reconstruction error of the GCN and Graphormer modules, and the contrastive learning loss, our model demonstrates excellent performance across multiple datasets. This ensures high reconstruction quality while enhancing the clustering performance of node representations.

Our model has been validated through experiments on six datasets, demonstrating its superiority in clustering quality and its effectiveness and applicability in handling complex real-world network data. The experimental results indicate that GCL-GCN outperforms existing advanced methods in several evaluation metrics, significantly improving clustering accuracy, stability, and robustness. Specifically, our model excels in handling sparse and heterogeneous graph data, capturing both global and local relationships of nodes, and achieving efficient feature representation learning. These results not only validate the theoretical advantages of our approach but also showcase its broad potential and value in practical applications. With further optimization and extension, GCL-GCN is poised to play a crucial role in various fields, providing a powerful tool for the analysis and understanding of complex graph data.

The structure of this paper is as follows: **Section 1** provides an overview of the importance of deep clustering research, highlights the limitations of existing methods, and introduces the motivation and contributions of this study. **Section 2** offers a detailed review of existing deep graph clustering methods, identifying their limitations. **Section 3** presents the components and methodology of the GCL-GCN model. **Section 4** focuses on the experimental part of the model. **Section 5** discusses the potential limitations of the GCL-GCN model. **Section 6** summarizes the main contributions of this paper and outlines future research directions.

## 2. Related Work

## 2.1 Attribute Graph Clustering

Traditional graph clustering methods, such as KMEANS, perform clustering analysis by converting graph data into vector representations [11]. However, these methods face challenges of high computational complexity and poor performance in handling noise when dealing with large-scale graph data. To overcome these limitations, new methods combining deep learning and clustering techniques have emerged in recent years. These new methods leverage deep learning models' ability to automatically extract high-dimensional feature representations, thereby excelling in handling complex and high-dimensional data and enabling more effective clustering analysis. For instance, Xie et al. proposed Deep Embedded Clustering (DEC), which simultaneously learns feature representations and clustering assignments through deep neural networks, significantly improving the clustering performance of graph data [12]. Additionally, Shao et al. introduced the Deep Discriminative Clustering Network (DCN), which uses a convolutional autoencoder and a Softmax layer to predict clustering assignments and combines discriminative loss as embedding regularization to enhance clustering quality in the subspace [13]. Furthermore, Jiang et al. proposed Variational Deep Embedding (VaDE), which, within the framework of variational autoencoders, utilizes a Gaussian mixture model and deep neural networks to optimize the clustering process, demonstrating excellent clustering performance on multiple benchmark datasets [14].

## 2.2 Graph Convolutional Network-based Clustering

GCNs have been widely applied in deep clustering due to their powerful ability to represent graph structures and node information, focusing on the structural information within graph data. For instance, the Embedded Graph Autoencoder (EGAE) model designed an embedded graph autoencoder by simultaneously learning relaxed KMEANS and GAE, inducing neural networks to generate deep features suitable for specific clustering models, making the representations applicable not only for graph clustering but also for other tasks [15]. The Structure Deep Clustering Network (SDCN) transfers the representations of autoencoders to the GCN layer through a transition operator and dual self-supervised mechanisms, achieving a natural combination of low-order to high-order structures and multiple representations. The Cross Attention Enhanced Graph Convolutional Network (CaEGCN) proposed a cross-attention-based enhanced GCN framework, integrating a content autoencoder module and GCN through cross-attention, combined with a self-supervised model, improving performance and robustness across different datasets [16]. The Mutual Boost Network (MBN) aimed to obtain comprehensive yet distinctive feature representations by proposing a dual-channel network that fuses AE and GAE representations, integrating AE with GCN representations as an enhanced module to propagate and integrate structural and node information obtained from other networks. The Deep Fusion Clustering Network (DFCN) proposed a structure and attribute information fusion (SAIF) module based on mutual learning, achieving consistency learning by merging AE and GCN representations to enhance clustering performance [17].

## 2.3 Contrastive Learning-based Clustering

In the realm of GNNs and representation learning for graph-structured data, contrastive learning has also demonstrated outstanding performance. As a key technology that does not require manual labels, contrastive learning plays a vital role in feature representation learning. By maximizing the similarity of feature representations of similar samples and minimizing the similarity of feature representations of dissimilar samples, contrastive learning significantly enhances the discriminative power of feature representations. You et al. (2020) proposed the GraphCL framework, which designed four graph augmentation methods to generate graph representations with excellent generalization, transferability, and robustness [18]. Zhu et al. (2020) introduced a new unsupervised graph representation learning framework that utilizes contrastive objectives at the node level, significantly outperforming existing methods, especially in prediction tasks [19]. In the study of graph data representation learning, methods combining GCN encoders and joint contrastive losses have also shown great potential. Zhu et al. (2022) proposed the Structure-Enhanced Heterogeneous Graph Contrastive Learning (STENCIL) model, which effectively improved the model's performance in complex graph structures by generating multiple semantic views and explicitly using structural embeddings [20]. Contrastive learning in GNNs and graph-structured data representation learning has shown its remarkable advantages by effectively leveraging intrinsic supervisory information, maintaining data structure and attributes, and designing consistent objective functions [21].

## 2.4 Graph Transformer: Encoding & Clustering

The application of Transformers to graph data is an emerging research area. The Transformer model, initially proposed by Vaswani et al. (2017), has performed excellently in various tasks, with its core mechanism being the attention mechanism [7]. Peng et al. (2021) proposed the Attention-driven Graph Clustering Network (AGCN), which dynamically integrates node attribute features and topological graph features [22]. Ying et al. (2021) introduced the Graphormer model, based on the standard Transformer architecture, which effectively encodes structural information of graphs, demonstrating superior performance in graph representation learning tasks [3]. Dwivedi and Bresson (2021) presented the Graph Transformer, extending the Transformer network to arbitrary graphs by using a neighborhood-based attention mechanism and Laplacian eigenvectors for positional encoding [23]. Baek et al. (2021) introduced the Graph Multiset Transformer (GMT), which employs a global pooling layer with multi-head attention to capture structural dependencies between nodes [24]. Wang et al. (2022) proposed ClusterFormer, which improves the efficiency and effectiveness of the Transformer through a neural clustering attention mechanism. This method has shown excellent performance in machine translation, text classification, natural language inference, and text matching tasks [25].

## 2.5 Limitations

GCN-based methods have garnered attention due to their effectiveness in capturing graph structural information. However, many methods still have limitations in critical aspects. For example, GAE performs poorly when handling highly sparse and heterogeneous static graph data because it relies on the reconstruction of graph structures and node features [26]. On the other hand, GCNN may face information loss issues when dealing with complex graph structures, mainly because its clustering loss function focuses on node aggregation, neglecting fine-grained information within complex graph structures. To address these limitations, researchers have begun exploring Transformer-based clustering models. The core of the Transformer model is the attention mechanism, which has shown excellent performance in multiple tasks when applied to graph data. However, these methods also have some limitations. The Graph Transformer model faces difficulties when dealing with sparse graph data because the self-attention mechanism may lead to insufficient information processing in sparse connections [27]. Although GTN excels at capturing global dependencies, it falls short in capturing local structural information. Additionally, Transformer models are generally less effective than specially designed GCNs in handling local graph structures [28]. Apart from Transformer models, the application of contrastive learning in graph data clustering has also gained widespread attention. While contrastive learning plays a crucial role in feature representation learning, using shallow architectures may fail to capture deep graph structural information, thus affecting the understanding and representation of complex graph data [29]. In practice, it has been observed that using complex architectural designs for contrastive learning modules may lead to a decline in clustering performance, possibly due to the amplification of noise within the data, resulting in inaccurate learned features [30]. For data with highly specific or sparse characteristics, the over-complication of graph structures and node features may fail to generate meaningful contrastive views [31].

## 3. Methodology

## 3.1 Overall Framework

We have designed a dual self-supervised tri-channel graph fusion network structure. As illustrated in Figure 1, this architecture primarily comprises the contrastive learning module, AE module, GCN module, Graphormer module, representation enhancement module, and self-supervised module. Firstly, the input graph data's original feature matrix $\mathbf{X}$ is fused with the feature matrix $\mathbf{X_c}$ output by the contrastive learning module. This fused feature matrix is then passed through the GCN module and Graphormer module, respectively, generating representations $\mathbf{Z^{(4)}}$ and $\mathbf{T^{(4)}}$ at the final layer of the encoders in these two modules. The AE module solely processes the original features of the graph data, outputting the representation $\mathbf{H^{(4)}}$ at the final layer of its encoder. The output representations from each encoder undergo a fusion operation to generate a new distribution $Q$. Subsequently, a target distribution $P$ is obtained through soft assignment, and $P$ is utilized to guide the self-supervised clustering process. Our model achieves efficient representation and processing of graph-structured data, and optimizes the similarity between different distributions using the KL divergence.

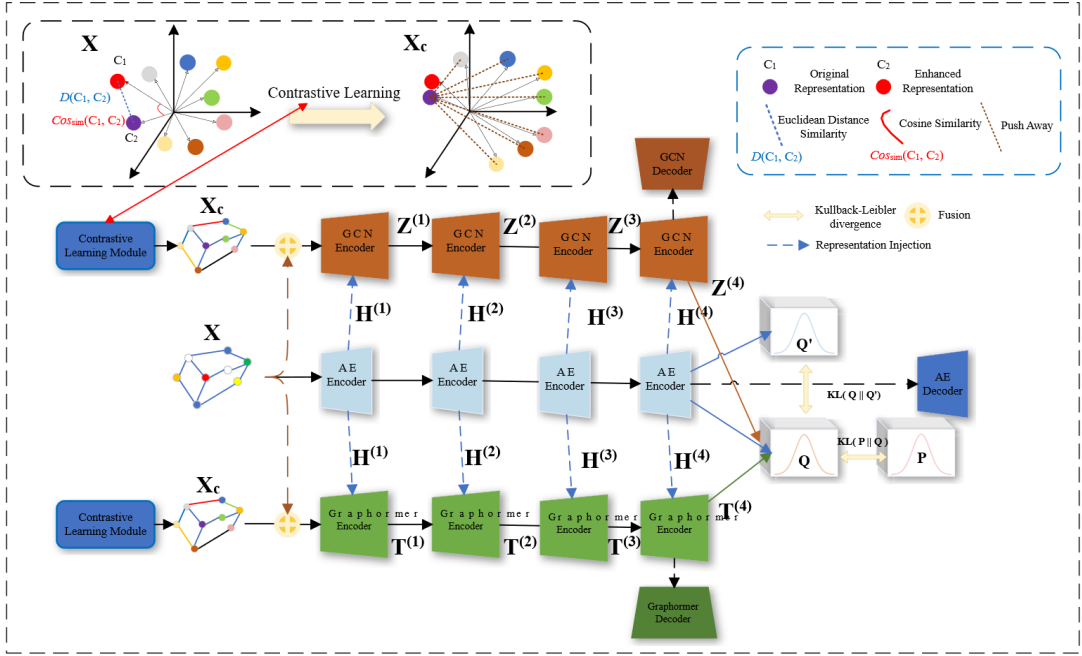

**Fig. 1.** The GCL-GCN Model Framework.

## 3.2 Auto Encoder Module

In our AE module, we employ a standard autoencoder structure composed of linear fully connected layers to extract representations from node attributes. The structure of the AE module is shown in Figure 2. Specifically, the representation of the $\ell$-th layer encoder in the AE can be expressed as:

$$\mathbf{H}^{(\ell+1)} = \sigma\left(\mathbf{W}_e^{(\ell+1)}\mathbf{H}^{(\ell)} + \mathbf{b}_e^{(\ell+1)}\right) \tag{1}$$

In Eq. (1), the linear transformation is performed by the weight matrix $\mathbf{W}_e^{(\ell+1)}$ and the bias vector $\mathbf{b}_e^{(\ell+1)}$, followed by a non-linear transformation through the activation function $\sigma$ (Leaky ReLU). During this process, the weight matrix $\mathbf{W}_e^{(\ell+1)}$ and the bias vector $\mathbf{b}_e^{(\ell+1)}$ are parameters that need to be trained. They are continuously updated during training through the backpropagation algorithm.

The representation of the AE's $\ell$-th layer decoder can be expressed as:

$$\hat{\mathbf{H}}^{(\ell+1)} = \sigma\left(\mathbf{W}_d^{(\ell+1)}\hat{\mathbf{H}}^{(\ell)} + \mathbf{b}_d^{(\ell+1)}\right) \tag{2}$$

In Eq. (2), $\hat{\mathbf{H}}^{(\ell+1)}$ represents the output of the decoding layer. $\mathbf{W}_d^{(\ell+1)}$ and $\mathbf{b}_d^{(\ell+1)}$ are the weight and bias parameters of this layer, respectively. These parameters are used to weight and shift the output of the previous layer $\hat{\mathbf{H}}^{(\ell)}$, and then the activation function $\sigma$ is applied to obtain the decoded result.

The formula for the loss function used to evaluate the prediction error of the model is described below:

$$\mathcal{L}_{res} = \frac{1}{2N} \sum_{i=1}^{N} \parallel x_i - \hat{x}_i \parallel_F^2 \tag{3}$$

In Eq. (3), the error is measured by calculating the square of the difference between the true value $x_i$ and the predicted value $\hat{x}_i$ for each sample, and then averaging the sum of the squared errors across all samples.

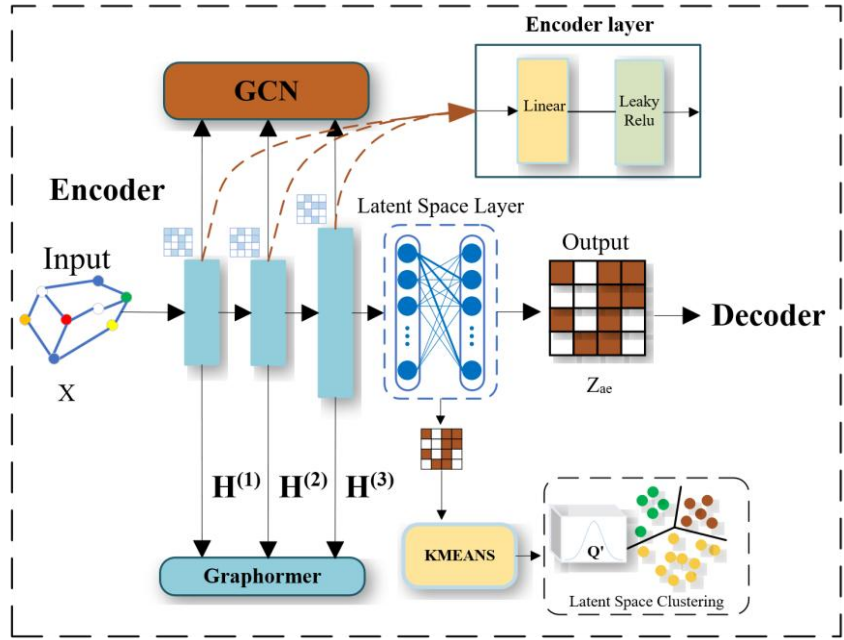

**Fig. 2.** The architecture of the AE module illustrates the feature matrix $\mathbf{X}$ as the input. The encoder section employs multiple layers of linear transformations and Leaky ReLU activation functions to convert the input features $\mathbf{X}$ into a low-dimensional representation. Subsequently, the output of each layer is processed by GCN and Graphormer layers to enhance the node information representation. In the final layer of the encoder, the low-dimensional representation is used as input for KMEANS clustering. The decoder section utilizes the same linear transformations and activation functions to restore the low-dimensional representation back to the original feature space, ultimately generating the reconstructed feature matrix $\mathbf{X}'$.

## 3.3 Graph Convolutional Network Module

As GCN can integrate node features with the topological structure of graph-structured data, we have designed a GCN module to enhance the model's capability to represent and understand graph data, as shown in Figure 3. In our framework, GCN effectively captures the structural and node feature information on the attributed graph, thereby generating richer and more accurate node embeddings. The propagation rule can be formally described as follows:

$$\mathbf{Z}_{GCN}^{(\ell+1)} = \sigma\left(\mathbf{E}\mathbf{Z}_{GCN}^{(\ell)}\mathbf{W}_e^{(\ell)}\right) \tag{4}$$

$\mathbf{Z}_{GCN}^{(\ell)}$ represents the node feature representation matrix at layer $\ell$-th, encompassing the feature vectors of each node in the current layer. These feature vectors undergo a graph convolution operation, which fuses the topological information of the graph with the initial features of the nodes, thereby capturing the deep relationships and attributes between nodes. $\mathbf{W}_e^{(\ell)}$ is the trainable weight matrix at layer $\ell$-th, used to perform linear transformations on the node features to update the node representations.

Specifically, we add the feature matrix $\mathbf{X}$ of the original data to the output feature matrix $\mathbf{X_c}$ from the contrastive learning module to form the input feature matrix for this layer. This, combined with the edge index matrix $\mathbf{E}$, serves as the input to the first layer of the GCN.

$$\mathbf{Z}_{GCN}^{(1)} = \sigma\left(\mathbf{E}(\mathbf{X} + \mathbf{X_c})\mathbf{W}_e^{(0)}\right) \tag{5}$$

After the graph convolution operation, the updated feature vector for each node in the graph is computed according to Eq. (6) as follows:

$$\mathbf{x}_i' = \Theta^\top \sum_{j \in \mathcal{N}(i) \cup i} \frac{e_{j,i}}{\sqrt{\hat{d}_j \hat{d}_i}} \mathbf{x}_j \tag{6}$$

The weight matrix $\Theta$ transforms the aggregated node feature vectors into a new feature space to better express complex patterns. The weight $e_{j,i}$ serves as the aggregation weight of the node features, reflecting the connection strength from node $j$ to node $i$. Here, the degree of node $i$, $\hat{d}_i$ (including self-loops), is defined as: $\hat{d}_i = 1 + \sum_{j \in \mathcal{N}(i)} e_{j,i}$, ensuring that the updated node features depend not only on the features of neighboring nodes but also on the features of the node itself. This enhances the stability and effectiveness of the graph convolution operation.

The representation of the GCN's $\ell$-th layer decoder can be expressed as:

$$\mathbf{Z}_{GCN}^{(\ell+1)} = \sigma\left(\mathbf{E}\mathbf{Z}_{GCN}^{(\ell)}\mathbf{W}_d^{(\ell)}\right) \tag{7}$$

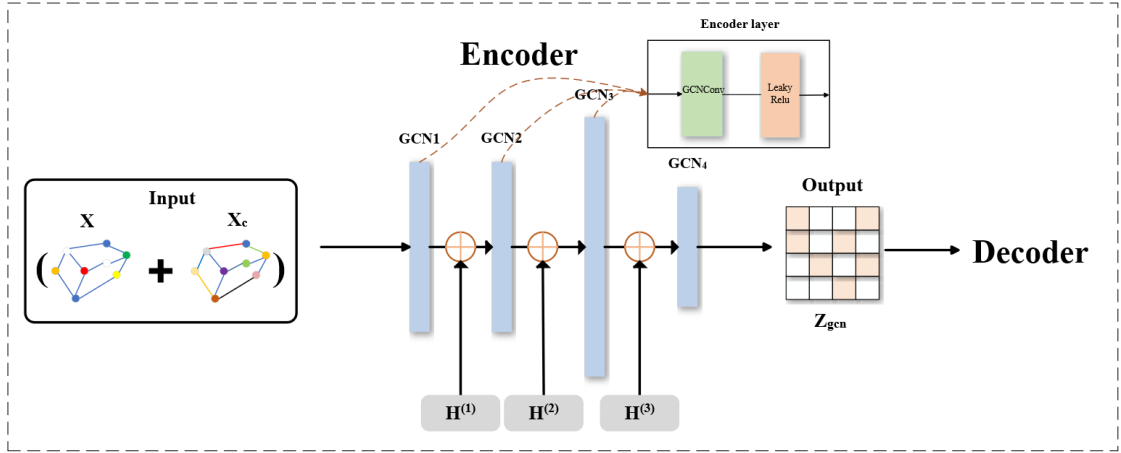

**Fig. 3.** The framework structure of the GCN module. First, the input graph data $\mathbf{X}$ and the feature matrix $\mathbf{X_c}$ generated by contrastive learning are combined to create new fused graph data. This fused data, along with the edge information, serves as the enhanced graph data input to the module. Subsequently, the output of each layer in the AE module is sequentially injected into the corresponding GCN layer. Finally, $\mathbf{Z}_{GCN}$ is used as the final output of the model for subsequent tasks.

## 3.4 Graphormer Module

The Graphormer module we designed is an innovative graph neural network layer that combines graph structure and node features for efficient graph data learning. This paper proposes a new centrality encoding method, using multiple centrality measures to construct a comprehensive centrality encoding scheme, allowing the module to capture node importance more thoroughly. This module enhances the representation capability of graph neural networks through Transformer convolution operations, excelling particularly in handling graphs with complex connection structures. The Graphormer module introduces centrality features and spatial relationships to improve the effectiveness of node feature aggregation and message passing, thereby enhancing the quality of graph representation. The structure of each layer of the Graphormer module is shown in Figure 4.

**Centrality Encoding**
Our Graphormer module leverages various centrality encoding methods to enhance node representation, including degree centrality, betweenness centrality, and closeness centrality. Degree centrality measures the number of connections a node has, and its normalized calculation formula is as follows:

$$C_D(v) = \frac{deg(v)}{max(deg(u))} \tag{8}$$

Where $deg(v)$ represents the degree of node $v$, and $max(deg(u))$ is the highest degree among all nodes in the graph. Betweenness centrality measures the frequency at which a node appears as an intermediary on all shortest paths. Its formula is given by:

$$C_B(v) = \sum_{s \neq v \neq t} \frac{\sigma_{st}(v)}{\sigma_{st}} \tag{9}$$

Where $\sigma_{st}(v)$ is the total number of shortest paths from node $s$ to node $t$, and $\sigma_{st}(v)$ is the number of those paths that pass through node $v$. Closeness centrality measures the average shortest path length from a node to all other nodes. Its calculation formula is:

$$C_C(v) = \frac{1}{\sum_{u \neq v} d(u,v)} \tag{10}$$

Where $d(u,v)$ represents the shortest path length from node $u$ to node $v$.

The comprehensive centrality combines degree centrality, betweenness centrality, and closeness centrality into a feature vector that encapsulates diverse information:

$$C(v) = [C_D(v), C_B(v), C_C(v)] \tag{11}$$

**Spatial Encoding**

To further enhance the aggregation of node features, the Graphormer module also calculates the spatial relationships between nodes. First, the Euclidean distance matrix of the node features is computed:

$$d(i,j) = \sqrt{\sum_{k=1}^{n} (x_{ik} - x_{jk})^2} \tag{12}$$

Where $x_{ik}$ and $x_{jk}$ are the values of node $i$ and node $j$ in the $k$-th feature dimension, respectively. Then, for each pair of connected nodes, their spatial relationship is calculated.

The core of the Graphormer module is the TransformerConv layer, which combines node features and centrality features, utilizing a multi-head attention mechanism for message passing. The primary operations of the TransformerConv layer are as follows:

Let the node features be $\mathbf{X} \in \mathbb{R}^{N \times F}$ and the centrality features be $\mathbf{C} \in \mathbb{R}^{N \times C}$, where $N$ is the number of nodes, $F$ is the dimension of the node features, and $C$ is the dimension of the centrality features. First, linear transformations are applied to the node features and centrality features:

$$\mathbf{H}_{\text{key}} = \mathbf{X}\mathbf{W}_{\text{key}} + \mathbf{C}\mathbf{W}_{c_{k}ey} \tag{13}$$

$$\mathbf{H}_{\text{query}} = \mathbf{X}\mathbf{W}_{\text{query}} + \mathbf{C}\mathbf{W}_{c_{query}} \tag{14}$$

$$\mathbf{H}_{\text{value}} = \mathbf{X}\mathbf{W}_{\text{value}} + \mathbf{C}\mathbf{W}_{c_{value}} \tag{15}$$

Where $\mathbf{W}_{\text{key}}, \mathbf{W}_{\text{query}}, \mathbf{W}_{\text{value}}, \mathbf{W}_{c_{k}ey}, \mathbf{W}_{c_{query}}$ and $\mathbf{W}_{c_{v}alue}$ are the learned weight matrices.

The feature matrices obtained from the linear transformations are then divided into multiple heads, and the attention scores for each head are calculated as follows:

$$\alpha_{ij}^{h} = \frac{\left(\mathbf{H}_{\text{query},i}^{h} \cdot \mathbf{H}_{\text{key},j}^{h}\right)}{\sqrt{d_k}} + d(i,j) \tag{16}$$

Where $\alpha_{ij}^{h}$ represents the attention score between node $i$ and node $j$ in the $h$-th head, $d_k$ is the scaling factor, and $d(i,j)$ denotes the spatial relationship between nodes $i$ and $j$.

The attention scores are normalized using the Softmax function, and the weighted sum is computed to obtain the new node feature representations:

$$\mathbf{T}_i^h = \sum_{j \in \mathcal{N}(i)} \alpha_{ij}^h \mathbf{H}_{\text{value},j}^h \tag{17}$$

Where $\mathcal{N}(i)$ represents the set of neighboring nodes of node $i$.

The output features from all heads are concatenated or averaged to obtain the final node feature representation:

$$\mathbf{T}_i = \frac{1}{H} \sum_{h=1}^{H} \mathbf{Z}_i^h \tag{18}$$

Here, $H$ denotes the number of attention heads.

Each layer of the Graphormer is composed of the TransformerConv operation with the new encoding method. The representation learned by the $\ell$-th layer of the encoder can be described as:

$$\mathbf{T}^{(\ell+1)} = \sigma\left(\mathbf{E}\mathbf{T}^{(\ell)}\mathbf{U}_e^{(\ell)}\right) \tag{19}$$

Where $\mathbf{U}_e^{(\ell)}$ is the weight matrix, $\mathbf{E}$ is the edge index matrix, and $\mathbf{T}^{(\ell)}$ is the output of the previous layer.

The representation of the $\ell$-th layer decoder in Graphormer can be described as:

$$\widehat{\mathbf{T}}^{(\ell+1)} = \sigma\left(\mathbf{E}\widehat{\mathbf{T}}^{(\ell)}\mathbf{U}_d^{(\ell)}\right) \tag{20}$$

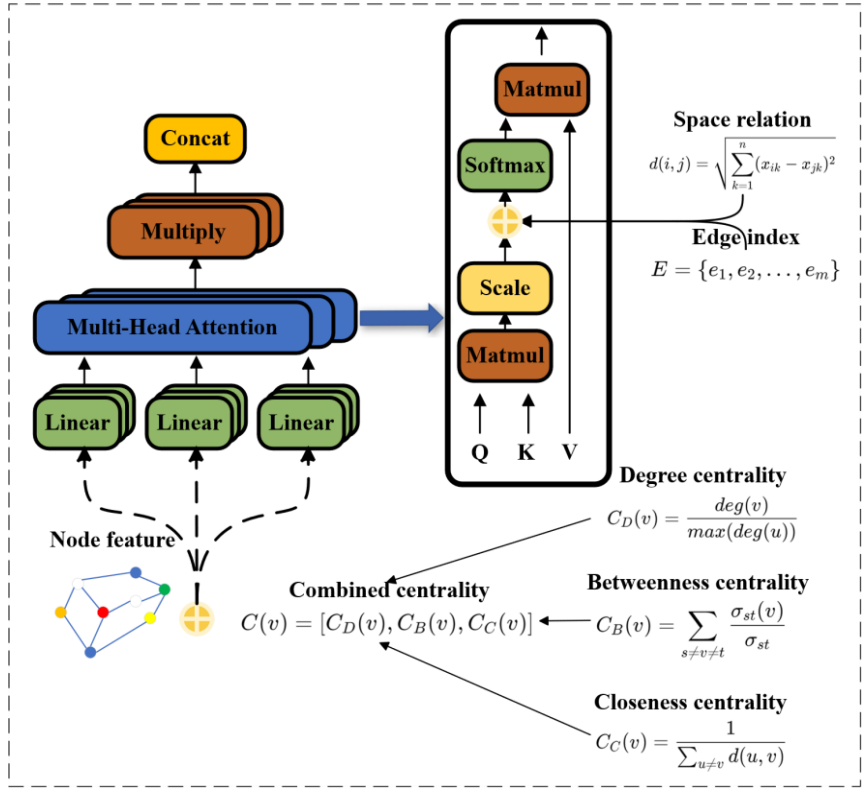

**Fig. 4.** illustrates the overall structure of the Graphormer module. Graphormer uses comprehensive centrality as an additional signal for the neural network. The comprehensive centrality encoding consists of three centrality measures for each node. Since comprehensive centrality encoding is applied to each node, node features can be combined with comprehensive centrality encoding as input. These features are then transformed linearly to generate keys, queries, and values, which are fed into the multi-head attention mechanism. The multi-head attention mechanism calculates attention weights between nodes, incorporating spatial relationships and edge indices. Through scaling, Softmax normalization, and weighted summation, the final node feature representations are generated.

## 3.5 Contrastive Learning Module

The design of the contrastive learning module aims to enhance the robustness of node representations by integrating cosine similarity and Euclidean similarity, where Euclidean similarity is derived from Euclidean distance. This approach, combined with data augmentation techniques, improves the model's adaptability to data variations. The module generates node representations of the original and augmented data through GCN, utilizes weighted parameters to combine the two similarity metrics, and optimizes the model using the cross-entropy loss function. This ensures that representations of similar nodes are closer together, while those of dissimilar nodes are further apart. The model encodes the original and augmented data to obtain embedding representations. These two embeddings, representing different views of the same data point, form positive sample pairs. When computing the contrastive learning loss, other samples within the batch are used as negative samples. Specifically, for each sample within a batch, the other samples in the batch are treated as negative samples. Additionally, through data augmentation with random masking and backpropagation optimization, the model parameters are continuously updated to minimize contrastive loss, further enhancing the accuracy and robustness of node representations. The structure of the contrastive learning module is shown in Figure 5.

Positive samples are generated through data augmentation. The original and augmented representations of the same node form positive sample pairs. Negative samples refer to the representations of other nodes. To improve the robustness of the model, we employ a simple data augmentation method by randomly dropping some features. Specifically, a mask is randomly generated to zero out parts of the node features, creating an augmented view. This augmented data helps the model better learn both the differences and commonalities in node representations. The data augmentation technique is formulated as follows:

$$\mathbf{X}_{aug} = \mathbf{X} \odot \mathbf{M} \tag{21}$$

Where $\odot$ denotes element-wise multiplication. $\mathbf{M}$ is a mask matrix with the same shape as $\mathbf{X}$, with elements $\mathbf{M}_{ij} \sim \text{Bernoulli}(1 - p)$, where $p$ is the dropout rate.

We first designed a graph neural network as an encoder to map node feature vectors into a low-dimensional embedding space. This encoder consists of two graph convolution layers. The first layer maps the input features to a hidden space, and the second layer further maps the hidden space to the output embedding space. The forward propagation process of the GCN module is as follows:

$$\mathbf{C}^{(1)} = \sigma\left(\widetilde{\mathbf{A}}\mathbf{X}\mathbf{W}^{(0)}\right) \tag{22}$$

$$\mathbf{C} = \widetilde{\mathbf{A}}\mathbf{C}^{(1)}\mathbf{W}^{(1)} \tag{23}$$

Where $\widetilde{\mathbf{A}} = \mathbf{D}^{-1/2}(\mathbf{A} + \mathbf{I})\mathbf{D}^{-1/2}$ is the adjacency matrix with added self-loops and symmetric normalization, $D$ is the degree matrix, $\sigma$ is the ReLU activation function, and $\mathbf{W}^{(0)}$ and $\mathbf{W}^{(1)}$ are the weight matrices.

To more accurately measure the similarity between node embeddings, we propose a hybrid similarity calculation method that combines cosine similarity and Euclidean similarity. Specifically, we first compute the cosine similarity between the embeddings of two nodes, then calculate their Euclidean distance and convert it into a similarity score. The final similarity is the weighted product of these two similarities, regulated by the parameter $\beta$.

$$Cos_{sim}(\mathbf{C}_1, \mathbf{C}_2) = \frac{\mathbf{C}_1\mathbf{C}_2^T}{\parallel \mathbf{C}_1 \parallel \parallel \mathbf{C}_2 \parallel} \tag{24}$$

$$\mathrm{Euc}_{\mathrm{Dist}}(\mathbf{C}_1, \mathbf{C}_2) = \sqrt{\sum_{i=1}^{n}(\mathbf{C}_{1,i} - \mathbf{C}_{2,i})^2} \tag{25}$$

$$\mathrm{Euc}_{\mathrm{sim}}(\mathbf{C}_1, \mathbf{C}_2) = \frac{1}{1 + \mathrm{Euc}_{\mathrm{Dist}}(\mathbf{C}_1, \mathbf{C}_2)} \tag{26}$$

$$\mathrm{Comb}_{\mathrm{sim}}(\mathbf{C}_1, \mathbf{C}_2) = (Cos_{sim}(\mathbf{C}_1, \mathbf{C}_2) \times \mathrm{Euc}_{\mathrm{sim}}(\mathbf{C}_1, \mathbf{C}_2))^{\beta} \tag{27}$$

Let $\mathbf{C}_1$ and $\mathbf{C}_2$ be two embedding matrices obtained through the GCN, representing the node feature representations. $T$ denotes the transpose operation. By converting the Euclidean distance into similarity (where greater distances correspond to lower similarity), we combine the similarities using the product of both cosine and Euclidean similarities, with the parameter $\beta$ controlling the balance between them.

In contrastive learning, we design a cross-entropy-based loss function to maximize the similarity between positive sample pairs and minimize the similarity between negative sample pairs. The similarity calculation and cross-entropy loss function in contrastive learning are formulated as follows:

$$\mathcal{L}_{\mathrm{contrastive}} = \frac{1}{N}\sum_{i=1}^{N} -\log\frac{e^{\mathrm{Comb}_{\mathrm{sim}}(\mathbf{c}_{1,i}, \mathbf{c}_{2,i})/\tau}}{\sum_{j=1}^{N} e^{\mathrm{Comb}_{\mathrm{sim}}(\mathbf{c}_{1,i}, \mathbf{c}_{2,i})/\tau}} \tag{28}$$

$N$ is the batch size. By combining the similarity matrix with the temperature parameter $\tau$, we maximize the prediction probability of the correct similarity. In Eq. (29), the numerator represents the similarity of the sample pair in an exponential form adjusted by the temperature parameter, while the denominator is a normalization term that includes the sum of similarities between the sample and all other samples.

After pre-training, we use the trained GCN model to compute the final node embeddings and save these representations for subsequent tasks. We use the original features as input to the pre-trained model to generate the final feature representations:

$$\mathbf{X}_c = \sigma(\mathbf{EXW}) \tag{29}$$

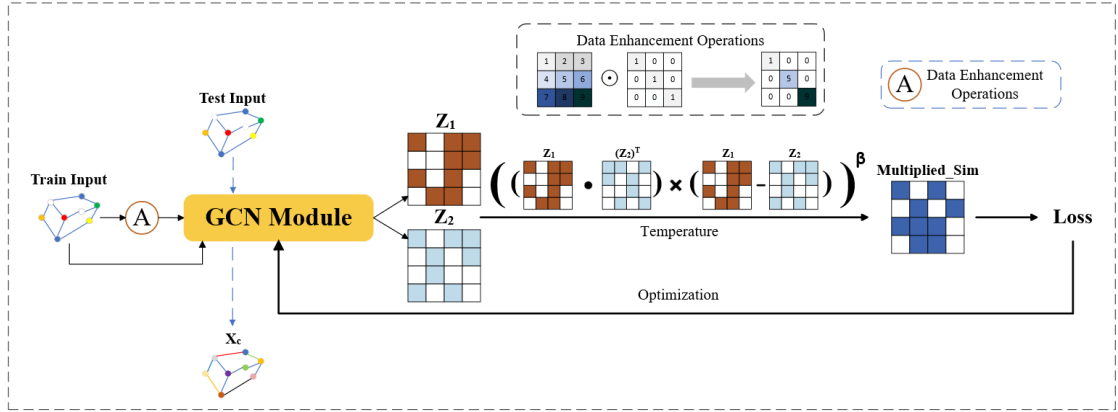

**Fig. 5.** The structure of the contrastive learning module encompasses the following steps: First, data augmentation operations (A) are performed on the training input data to generate augmented input. Next, both the augmented and original input data are fed into the GCN module, which extracts node feature representations through two GCNs, resulting in the feature matrices $\mathbf{C}_1$ and $\mathbf{C}_2$. Subsequently, the cosine similarity and Euclidean similarity between $\mathbf{C}_1$ and $\mathbf{C}_2$. are calculated and combined to obtain the multiplicative similarity matrix. This matrix is then integrated with the temperature parameter to compute the contrastive learning loss, which is optimized using the cross-entropy loss function. Finally, the Adam optimizer is employed to refine the model, updating its parameters to generate the final feature representation $\mathbf{X}_c$, and the model output is saved.

## 3.6 Dual Self-Supervised Module

Our framework employs a dual self-supervised module to guide network training, thereby generating more accurate cluster assignments. This strategy consists of two steps: (1) calculating the soft cluster assignments $\boldsymbol{Q}$ and $\boldsymbol{Q}'$, and (2) generating the target distribution $\boldsymbol{P}$ from $\boldsymbol{Q}$.

**Soft Cluster Assignment Calculation**

We use the node embeddings generated by the AE as the initial input and run the KMEANS algorithm multiple times randomly to ensure the selection of cluster centers is more robust and stable. We employ the Students' t-distribution, as proposed by Van der Maaten & Hinton (2008), to calculate the similarity between each cluster center $\boldsymbol{C}_j$ and node representation $\boldsymbol{Z}_j$ [32].

Additionally, we adopt the Sum of Squared Errors (SSE) as an internal validity metric, as proposed by Ünlü & Xanthopoulos (2019), to achieve more reasonable clustering results [33]. Thus, our soft cluster assignment $Q_{ij}$ is defined as Eq. (30):

$$q_{ij} = \frac{(1+\| z_i - c_j \|^2/t)^{-\frac{t+1}{2}}}{\sum_{j'} (1+\| z_i - c_{j'} \|^2/t)^{-\frac{t+1}{2}}} \tag{30}$$

**Target Distribution Generation**

To enhance the weights of high-confidence samples and reduce the impact of low-confidence samples, we define the target distribution shown in Equation (31). This allows the model to better focus on reliable samples during the training process.

$$p_{ij} = \frac{q_{ij}^2/f_j}{\sum_{j'} q_{ij'}^2/f_{j'}} \tag{31}$$

Where $P_i$ represents the target distribution probability of node $i$ in cluster $j$, and $Q_{ij}$ is the soft cluster assignment probability of node $i$ in cluster $j$. The parameter $f_j$ can be defined as $f_j = \sum_i q_{ij}$, representing the frequency of cluster $j$, which is the sum of the soft cluster assignments for cluster $j$, used to normalize the soft cluster assignments.

By minimizing the Kullback-Leibler (KL) divergence between $P$ and $Q$, our model continuously optimizes the clustering results during the iterative process, thereby achieving higher clustering quality. The objective function is defined as follows:

$$L_{clu} = KL(P \parallel Q) = \sum_i \sum_j p_{ij} log\, g \frac{p_{ij}}{q_{ij}} \tag{32}$$

This loss function not only enhances the impact of high-confidence assignments but also, through the self-supervised learning strategy, makes data representations closer to cluster centers, thus improving the model's clustering performance.

Our framework's dual self-supervised module optimizes feature representations and improves model performance by integrating features generated by the GCN module, AE module, and Graphormer module, and dynamically adjusting feature representations and distributions through KL divergence.

Specifically, the GCN module captures local structural information in the graph data, while the Graphormer module excels at extracting global features from the graph data. The GCN and Graphormer modules further refine the feature representations using the outputs from each layer of the AE module, enabling the model to better adapt to complex graph data structures. We generate the soft cluster assignment distribution $\mathbf{Q}$ by integrating $\mathbf{Z_{AE}}$, $\mathbf{Z_{GCN}}$, and $\mathbf{Z_T}$, while the clustering distribution $\mathbf{Q'}$ is generated directly from $\mathbf{Z_{AE}}$. We quantify the difference between distributions $\mathbf{Q}$ and $\mathbf{Q'}$ using KL divergence, as shown in Equation (33).

$$L_{con} = KL(\mathbf{Q} \parallel \mathbf{Q'}) = \sum_i \sum_j q_{ij} log \frac{q_{ij}}{q'_{ij}} \tag{33}$$

By minimizing the KL divergence, we effectively align the generated feature distribution with the intrinsic structure of the data, thereby significantly enhancing our model's learning efficacy and generalization capability.

## 3.7 Representation Enhancement Module

In our model, we use an enhancement module to better leverage node attributes and the graph's topological structure, extracting more useful features to improve the model's performance. By integrating the latent representations generated by the AE module into each layer of the GCN and Graphormer modules, we ensure that the enhanced feature and structural information is effectively propagated throughout the different layers of the model, further improving its representation capability. The feature update formulas in the $\boldsymbol{\ell}$-th layer of the GCN and Graphormer modules are as follows:

$$\mathbf{Z}_{GCN}^{(\ell+1)} = \mathbf{GCN}\left(\varepsilon \mathbf{H}^{(\ell)} + (1-\varepsilon)\mathbf{Z}_{GCN}^{(\ell)}\right) \tag{34}$$

$$\mathbf{Z}_T^{(\ell+1)} = \mathbf{Graphormer}\left(\varepsilon \mathbf{H}^{(\ell)} + (1-\varepsilon)\mathbf{Z}_T^{(\ell)}\right) \tag{35}$$

The feature representation $\mathbf{Z}_{GCN}^{(\ell+1)}$ of the $(\boldsymbol{\ell}+1)$-th layer is obtained by linearly combining the convolution output $\mathbf{H}^{(\ell)}$ of the $\boldsymbol{\ell}$-th layer and the features $\mathbf{Z}_{GCN}^{(\ell)}$ from the previous layer, with weights $\varepsilon$ and $(1-\varepsilon)$, respectively. The result of this linear combination is then processed through the GCN module or Graphormer module. The formulas demonstrate how the feature representations are updated by weighted fusion of the current layer output and the previous layer features in both the GCN and Graphormer modules, enabling effective handling and representation learning of graph data.

The final feature representation $\mathbf{Z}_L$ is obtained by integrating the feature representations from the GCN, AE, and Graphormer modules. The formula is as follows:

$$\mathbf{Z}_L = \widetilde{\mathbf{A}} \left( \boldsymbol{\lambda} \cdot \mathbf{Z}_{GCN}^{(L)} + \boldsymbol{\theta} \cdot \mathbf{Z}_{AE}^{(L)} + \boldsymbol{\gamma} \cdot \mathbf{Z}_{T}^{(L)} \right) \tag{36}$$

The feature representation of the $\boldsymbol{\ell}$-th layer is generated by the last layer of the encoders from the GCN module, AE module, and Graphormer module, denoted as $\mathbf{Z}_{GCN}^{(L)}$, $\mathbf{Z}_{AE}^{(L)}$, and $\mathbf{Z}_{T}^{(L)}$ respectively. These feature representations are obtained by processing the input data through their respective modules. The parameters $\boldsymbol{\lambda}$, $\boldsymbol{\theta}$, and $\boldsymbol{\gamma}$ are weights that control the contribution of each feature representation in the final integration. By linearly combining these three weighted feature representations, the final integrated feature representation $\mathbf{Z}_L$ is obtained. This process leverages the advantages of each module, enhancing the overall representation capability and performance of the model.

## 3.8 Optimizer

The loss function of the model integrates the feature matrix reconstruction loss and the adjacency matrix reconstruction loss, taking into account the reconstruction errors of both node embeddings and edge embeddings. The overall loss for the GCN module and the Graphormer module is given as follows:

$$L_{\mathrm{G}} = L_w + 0.1 \cdot (L_{a1} + L_{a2}) \tag{37}$$

In Eq. (36), referring to the MBN model setup, the adjacency matrix reconstruction losses ($L_{a1}$ and $L_{a2}$) are multiplied by 0.1. This coefficient is used to balance the relative importance of the feature matrix reconstruction loss and the adjacency matrix reconstruction loss.

The model measures the error formula of the AE module in reconstructing the input feature matrix as follows:

$$L_w = \frac{1}{N} \sum_{i=1}^{N} \left( \frac{1}{2} \hat{Z}_{GCN,i} + \frac{1}{2} \hat{Z}_{T,i} - (AX)_i \right)^2 \tag{38}$$

Where $\hat{Z}_{GCN,i}$ and $\hat{Z}_{T,i}$ are the weighted average node embeddings generated by the GCN module and the Graphormer module, respectively. These embeddings are then compared with the product of the adjacency matrix $A$ and the input feature matrix $\mathbf{X}$ by calculating the mean squared error. By minimizing this loss, our model can better learn the node embeddings, ensuring that the reconstructed feature matrix is as close as possible to the original input feature matrix.

The adjacency matrix reconstruction loss is as follows:

$$L_{a1} = \frac{1}{N^2} \sum_{i=1}^{N} \sum_{j=1}^{N} \left( \widehat{\mathbf{A}}_{GCN,ij} - \mathbf{A}_{ij} \right)^2 \qquad (39)$$

$$L_{a2} = \frac{1}{N^2} \sum_{i=1}^{N} \sum_{j=1}^{N} \left( \widehat{\mathbf{A}}_{T,ij} - \mathbf{A}_{ij} \right)^2 \qquad (40)$$

These equations calculate the mean squared error between the predicted adjacency matrices generated by the GCN module and the Graphormer module ($\hat{A}_{\text{GCN},ij}, \hat{A}_{\text{T},ij}$) and the actual adjacency matrix. This measures the model's performance in reconstructing the graph structure. By minimizing this loss, our model can better capture the structural information of the graph, ensuring that the predicted adjacency matrices are as close as possible to the actual adjacency matrix.

The mean squared error for the AE module in reconstructing the input feature matrix is as follows:

$$L_{AE} = \frac{1}{N} \sum_{i=1}^{N} \left( \widehat{\mathbf{H}_{Gt}} - (\widetilde{\mathbf{A}}\mathbf{X})_i \right)^2$$
$$\widehat{\mathbf{H}_{Gt}} = \widehat{\mathbf{H}_{AE}} \qquad (41)$$

This equation calculates the mean squared error between the feature matrix $\widehat{\mathbf{H}_{Gt}}$ reconstructed by the AE module and the input feature matrix $\mathbf{X}$ propagated through the adjacency matrix $\widetilde{\mathbf{A}}$, averaged over all nodes.

The total loss function $L$ consists of three parts: the reconstruction loss $L_{rec}$, the clustering loss $L_{clu}$, and the consistency loss $L_{con}$.

$$L = L_{rec} + \alpha L_{clu} + \beta L_{con}$$
$$L_{rec} = L_G + L_{AE} \qquad (42)$$

The reconstruction loss $L_{rec}$ includes the graph reconstruction loss $L_G$ and the AE module loss $L_{AE}$. The clustering loss and consistency loss contributions to the overall loss function are balanced using the weighting coefficients $\alpha$ and $\beta$.

The clustering loss $L_{clu}$ with respect to the cluster centers $c_j$ is optimized using the following gradient formula:

$$\frac{\partial L_{clu}}{\partial c_j} = -\frac{t+1}{t} \sum_{i=1}^{N} \left( 1 + \frac{\| z_i - c_j \|^2}{t} \right)^{-1} \times (p_{ij} - q_{ij})(z_i - c_j) \qquad (43)$$

The equation calculates the partial derivative of the clustering loss with respect to each cluster center, where $t$ is a scaling parameter, and $\| z_i - c_j \|$ represents the Euclidean distance between sample $z_i$ and cluster center $c_j$. The term $(1 + \frac{\| z_i - c_j \|^2}{t})^{-1}$ is the weight of the distance, and $p_{ij}$ and $q_{ij}$ are the soft assignment probability and the target distribution probability of sample $i$ belonging to cluster center $j$, respectively. By calculating the weighted difference between samples and cluster centers and adjusting the positions of the cluster centers, the model can better capture the clustering structure of the data.

The equation for assigning the cluster label $\mathbf{r}_i$ to node $i$ is as follows:

$$\mathbf{r}_i = ar\,g\max_j q_{ij} \tag{44}$$

Where $q_{ij}$ represents the probability that node $i$ belongs to cluster $j$. This formula identifies the value of $j$ that maximizes $q_{ij}$, thereby determining the cluster label $\mathbf{r}_i$ for node $i$. Using this formula, each node can be assigned to the cluster with the highest probability, thus completing the clustering task.

Algorithm 1 shows the procedure of the GCL-GCN.

| **Algorithm 1:** Graphormer and Contrastive Learning Enhanced GCN |
| --- |
| **Input**: Graph data (**X**, **Y**, **adj**, **edges**); Hyperparameters $\boldsymbol{\alpha}$, $\boldsymbol{\beta}$, $\boldsymbol{\lambda}$, $\boldsymbol{\theta}$, $\boldsymbol{\gamma}$; Iteration number $\mathbf{N_{Iter}}$; Cluster number **k**; |
| **Output**: Cluster partition results **r**; |
| 1   Initialize the parameters of auto-encoder with pre-training; |
| 2   Initialize the parameters of contrastive learning with pre-training; |
| 3   Initialize the weight matrix of GCN; |
| 4   Initialize the weight matrix of Graphormer; |
| 5   Initialize **k** cluster centroids based on the representation from pre-trained auto-encoder; |
| 6   **for** iter $\in$ 0, 1, $\cdots N_{Iter}$ **do** |
| 7       Obtain the representation $\boldsymbol{X_c}$ by Eq. (29); |
| 8       Obtain the representations $\boldsymbol{Z^{(l)}}$, $\boldsymbol{H^{(l)}}$ and $\boldsymbol{T^{(l)}}$ by Eq. (2), Eq. (4) and Eq. (19); |
| 9       Generate the enhanced representation $\mathbf{Z_L}$ by Eq. (36); |
| 10      Calculate the soft assignments $\mathbf{Q}$, $\mathbf{Q'}$ and $\mathbf{P}$ by Eq. (30) & Eq. (31); |
| 11      Feed the $\mathbf{Z}$, $\mathbf{H}$ and $\mathbf{T}$ to the decoder to reconstruct the raw data $\mathbf{X}$, $\mathbf{A}$; |
| 12      Calculate the objective function by Eq. (42); |
| 13      Optimize and update parameters of the whole model by back propagation; |
| 14  **end** |
| 15  Obtain the results of cluster partition **r** by Eq. (44); |

### 3.9 Complexity Analysis

The time complexity of our model is primarily determined by the AE module, the GCN module, and the Graphormer module. The AE module comprises three encoding layers, one bottleneck layer, and four decoding layers. Suppose the AE module has $L_{ae}$ layers, including the encoding layers, the bottleneck layer, and the decoding layers. The total time complexity of the AE module is:

$$O\left(N\sum_{l=1}^{L_{ae}} D_{in}^{(l)} \cdot D_{out}^{(l)}\right)$$ where $N$ is the number of nodes, and $D_{in}^{(l)}$ and $D_{out}^{(l)}$ are the input and

output feature dimensions of the $l$-th layer, respectively. For the GCN module, the computation mainly involves linear transformation and feature aggregation. In the linear transformation, the weight matrix is applied to each node, resulting in a complexity of: $O(ND_{in}D_{out})$ where $N$ is the number of nodes, $D_{in}$ is the input feature dimension, and $D_{out}$ is the output feature dimension. In the feature aggregation part, each node collects features from all its neighboring nodes, including itself, with a complexity of: $O(ED_{out})$ where $E$ is the number of edges in the graph. Assuming the input and output feature dimensions are approximately the same across all layers, i.e., $D_{in} \approx D_{out} \approx D$, and there are $L_{gcn}$ layers, the total time complexity of the GCN module is: $O(L_{gcn}(ND^2 + ED))$ The time complexity of the Graphormer module mainly depends on the TransformerConv operations in each layer. Suppose the Graphormer module has $L_g$ layers, the time complexity includes: Linear transformation: The complexity per head is $O(ND^2)$, and the total complexity is $O(ND^2H)$; Attention computation: The complexity per head is $O(ED)$, and the total complexity is $O(EDH)$; Weighted aggregation based on attention scores: The complexity is $O(EDH)$. where $N$ is the number of nodes, $D$ is the feature dimension per node, $E$ is the number of edges, and $H$ is the number of heads in multi-head attention. Therefore, the total time complexity of the Graphormer module is: $O(L_g(ND^2H + EDH))$ Assuming the AE layers, GCN layers, and Graphormer layers have $L_{ae}, L_{gcn}$ and $L_g$ layers respectively, the overall time complexity is:

$$O\left(\left(\sum_{l=1}^{L_{ae}} D_{in}^{(l)} \cdot D_{out}^{(l)}\right)N + L_{gcn}(ND^2 + ED) + L_g(ND^2H + EDH)\right).$$

## 4. Experiments

### 4.1 Benchmark Datasets

**Table 1.** The description of the datasets.

| Dataset | Type | Sample | Classes | Dimension |
| --- | --- | --- | --- | --- |
| HHAR | Record | 10299 | 6 | 561 |
| Reuters | Text | 10000 | 4 | 2000 |
| ACM | Graph | 3025 | 3 | 1870 |
| DBLP | Graph | 4057 | 4 | 334 |
| CiteSeer | Graph | 3327 | 6 | 3703 |
| Cora | Graph | 2708 | 7 | 1433 |

In our experiments, we employed six benchmark datasets to validate the effectiveness of our model. These datasets include graph datasets ACM[1] [34], DBLP [35], CiteSeer [36], Cora [37]; and non-graph datasets HHAR [38] and Reuters [39]. The attributes of these datasets are shown in Table 1. Below is a detailed description of each:

- ACM: This dataset comprises 3,025 papers categorized into three fields: databases, wireless communication, and data mining. The features of the papers correspond to elements of a bag-of-words model represented by keywords.

- DBLP: This dataset includes records of over 5.4 million papers, encompassing conference proceedings, journal articles, book chapters, etc. It is divided into four categories, with each research field of the authors considered a feature.

- CiteSeer: This dataset contains 3,327 scientific publications classified into six categories. It includes sparse word and feature vectors for each document, as well as a list of citation links between documents.

- Cora: This dataset comprises 2,708 scientific papers categorized into seven classes. Features include the title, authors, abstract, citation information, and classification labels of each paper. The papers are represented by binary word vectors indicating the presence of corresponding dictionary words in the papers.

- HHAR: This dataset consists of 10,299 sensor records used for human activity recognition and health monitoring with wearable devices. The samples are divided into six classes, reflecting the anticipated perceptual heterogeneity in real-world deployment by collecting human activity data.

- Reuters: This dataset contains 21,578 news documents from Reuters. Due to computational constraints, we randomly selected a subset of 10,000 samples. We chose four categories as labels: corporate/industrial, government/social, markets, and economics.

[1]*https://dl.acm.org/*

## 4.2 Evaluation criteria

We adopted four widely recognized and frequently used evaluation metrics: Accuracy (ACC), Normalized Mutual Information (NMI) [40], Adjusted Rand Index (ARI) [41], and F1-score [42]. These evaluation metrics collectively form a comprehensive comparison framework, enabling us to quantitatively and multidimensionally compare the performance of the models. Specifically, higher scores in these evaluation metrics indicate superior model performance.

## 4.3 Comparison methods

**AE.** A deep clustering method combining AE and KMEANS algorithms effectively utilizes the low-dimensional representations extracted by the autoencoder for clustering. The autoencoder reduces the dimensionality of the data, extracting key features, while the KMEANS algorithm performs clustering in this low-dimensional feature space, significantly improving clustering effectiveness [43].

**MBN.** MBN is a dual-channel network for attributed graph clustering, integrating AE and GAE to enhance clustering performance through mutual learning and interaction. This algorithm includes a representation enhancement module, injecting node representations generated by the AE module into the GAE module, layer by layer, to merge node and structural feature information, creating more comprehensive and discriminative representations [1].

**TDCN.** Transformer-based Dynamic Fusion Clustering Network (TDCN) is a deep clustering algorithm combining Transformer and AE. It integrates features extracted by both through a dynamic attention mechanism and uses transformation operation G to obtain latent structural information. Trained with a dual self-supervised mechanism, this design significantly enhances the capability to handle complex data structures and achieve excellent clustering results [2].

**DFCN.** DFCN is a deep learning clustering technique that integrates autoencoders and graph neural networks to enhance clustering performance. Its core is the SAIF module, which uses interdependence learning to explicitly integrate representations from both autoencoders and graph autoencoders to achieve consensus representation. DFCN also introduces a triple self-supervised strategy and a symmetric graph autoencoder (IGAE) to enhance the model's generalization performance [17].

**SDCN.** Structured Deep Clustering Network (SDCN) combines deep neural networks (DNN) and GCN for clustering. This algorithm uses an autoencoder module and a graph convolutional network module, guided by a dual self-supervised mechanism, to extract node features and graph structure information, ultimately obtaining more comprehensive and effective graph representations [44].

**SCGC.** Self-Supervised Contrastive Graph Clustering (SCGC) is a deep clustering method that incorporates graph structure information into node representations through self-supervised contrastive learning. This method uses neighbor-guided contrastive loss to make representations of connected nodes closer and iteratively optimizes node embeddings and clustering results [45].

**CONVERT.** CONTRASTIVE Graph Clustering network with Reliable AugmenTation (CONVERT) is an algorithm for graph contrastive clustering. It designs a reversible perturbation-recovery network to generate augmented views with reliable semantics and uses a label matching mechanism to leverage high-confidence pseudo-labels for clustering, effectively addressing the semantic drift problem of augmented views and significantly improving clustering performance [46].

**IDEC.** The IDEC algorithm enhances clustering performance by preserving local data structures while learning deep feature representations. This method addresses limitations of previous deep clustering algorithms, particularly issues with feature space disruption due to clustering loss. IDEC uses autoencoders to preserve the local structure of the data and clustering loss to guide the dispersion of data points in the embedding space, achieving more efficient clustering [47].

**GAE & VGAE.** GAE and VGAE are two efficient graph embedding methods. GAE learns low-dimensional node embeddings directly through GCN, simplifying the variational inference process by focusing on optimizing reconstruction loss via gradient descent, suitable for fast embedding learning in large-scale graph data. VGAE combines VAE and GCN to learn latent representations of undirected graphs using latent variables, encoding with GCN, and reconstructing the adjacency matrix through an inner product decoder for efficient graph data modeling and analysis [5].

**AGC.** AGC is a clustering method for attributed graphs. It captures the global clustering structure through high-order graph convolutions and adaptively selects the appropriate convolution order to optimize clustering performance for different graphs. AGC first applies high-order graph convolutions to node features to make adjacent node features more similar, then uses spectral clustering on the smoothed features [48].

**DAEGC.** This method combines deep learning with graph attention mechanisms, proposing an autoencoder based on attention mechanisms to improve clustering accuracy and efficiency by learning latent representations of graph structure and content information, enhancing modeling and analysis capabilities for complex graph data [42].

**ARGA.** Adversarially Regularized Graph Autoencoder (ARGA) improves graph embedding by combining graph convolutional autoencoders with adversarial regularization. This method embeds graph structure and node content into latent representations through an encoder and uses a discriminator for adversarial training to align latent representations with prior distributions, producing robust graph embeddings [49].

**EGAE.** EGAE combines the advantages of GAE and relaxed k-means theory to achieve efficient unsupervised graph node clustering. The algorithm optimizes the layout of graph representations in the inner product space, significantly enhancing clustering performance. Its advantage lies in simultaneously extracting node features and understanding the global structure of the graph, capturing intrinsic node attributes while considering the overall graph layout, thereby achieving higher clustering accuracy [15].

## 4.4 Implementation details

**Table 2.** Model parameter configurations for the 6 benchmark datasets.

| Dataset | epoch | $\alpha$ | $\beta$ | $n_z$ | Learning rate | $\lambda$ | $\theta$ | $\gamma$ | $\varepsilon$ |
|---------|-------|----------|---------|-------|---------------|-----------|----------|----------|---------------|
| ACM | 200 | 0.3 | 0.3 | 10 | 5e-5 | 0.4 | 0.3 | 0.3 | 0.5 |
| DBLP | 200 | 0.08 | 0.3 | 10 | 2e-3 | 0.7 | 0.1 | 0.2 | 0.5 |
| CiteSeer | 200 | 0.3 | 0.12 | 10 | 4e-5 | 0.1 | 0.8 | 0.1 | 0.5 |
| Cora | 400 | 0.1 | 0.1 | 10 | 1e-4 | 0.4 | 0.1 | 0.5 | 0.5 |
| HHAR | 600 | 0.15 | 0.05 | 20 | 1e-4 | 0.1 | 0.8 | 0.1 | 0.5 |
| Reuters | 200 | 0.3 | 0.3 | 20 | 1e-4 | 0.4 | 0.1 | 0.5 | 0.5 |

Prior to training the model, we performed 50 pre-training iterations on the AE and retained the weight parameters from the AE training. In the AE module, GCN module, and Graphormer module, we set the dimensions of each layer to 500, 500, 2000, and $n_z$. Specifically, the $n_z$ settings, the number of iterations, and the learning rates for each dataset are shown in Table 2. We used KMEANS to cluster the feature representations output by the last layer of the autoencoder, repeating the process 20 times to select the best result. We evaluated the clustering performance on six datasets using four evaluation metrics. To ensure the reproducibility of the experimental data, we set a random seed. We fixed the weighting parameter $\varepsilon$ for merging the representation layer of the AE module with the GCN module and the Graphormer module at 0.5. For the six datasets, we varied the weighting parameters $\lambda$, $\theta$, and $\gamma$ in the enhancement module among the GCN module, AE module, and Graphormer module according to the different datasets, as detailed in Table 1. The proposed GCL-GCN was implemented on a machine equipped with an Intel i5-1240P CPU, an NVIDIA GTX 3070 GPU, and 32 GB of memory. The deployment environment was a Windows 10 system running the PyTorch 3.9.0 platform.

## 4.5 Clustering Results

The clustering performance of GCL-GCN and the comparative algorithms is shown in Tables 3 and 4. Our model, through the AE module, GCN module, Graphormer module, and contrastive learning module, fully leverages the relationship between structural information and node information in graph data, significantly enhancing clustering accuracy. Experimental results demonstrate that GCL-GCN surpasses traditional clustering models and graph neural network-based clustering models across multiple metrics, particularly when handling complex relationships and highly nonlinear structures in data. Additionally, GCL-GCN also excels in comparisons with Transformer-based clustering models and contrastive learning-based clustering models.

**Table 3.** Clustering results on seven datasets (mean ± std) – 1

| Dataset | Metric | AE | GAE | VGAE | EGAE | DAEGC | ARGA | AGC | OURS |
|---------|--------|-----|------|------|------|-------|------|-----|------|
| ACM | ACC | 82.35±0.08 | 84.51±1.42 | 84.11±0.21 | 83.63±3.97 | <u>85.28±0.27</u> | 64.65±1.93 | 79.64±0.61 | **93.39±0.29** |
| | NMI | 48.73±0.16 | <u>55.39±1.91</u> | 53.20±0.51 | 52.00±7.92 | 55.01±0.83 | 30.75±2.10 | 48.82±0.93 | **76.45±0.08** |
| | ARI | 54.85±0.16 | 59.48±3.10 | 57.71±0.69 | 57.83±8.71 | <u>61.20±0.64</u> | 20.20±3.47 | 50.59±1.05 | **81.44±0.07** |
| | F1 | 82.19±0.08 | 84.68±1.31 | 84.19±0.22 | 34.22±26.63 | <u>85.31±0.28</u> | 64.01±1.99 | 80.08±0.59 | **93.39±0.03** |
| Cora | ACC | 49.93±0.17 | 63.80±1.29 | 64.34±1.34 | <u>68.97±3.68</u> | 67.72±0.19 | 59.44±4.50 | 64.90±2.35 | **73.24±0.06** |
| | NMI | 26.59±0.23 | 47.64±0.37 | 48.57±1.09 | 50.27±3.06 | 49.98±0.11 | 41.02±2.22 | <u>51.67±1.56</u> | **55.16±0.14** |
| | ARI | 21.94±0.16 | 38.00±1.19 | 40.35±1.46 | <u>44.61±4.04</u> | 43.39±0.19 | 34.27±3.29 | 41.05±3.42 | **52.19±0.11** |
| | F1 | 48.38±0.24 | 65.86±0.69 | 64.03±1.34 | 15.73±9.79 | <u>66.12±0.18</u> | 59.57±4.16 | **68.63±4.14** | 64.79±0.04 |
| CiteSeer | ACC | 62.37±0.13 | 61.38±0.80 | 60.98±0.38 | 52.83±5.12 | 66.55±0.11 | 42.65±3.27 | <u>67.87±0.10</u> | **71.37±0.34** |
| | NMI | 35.50±0.08 | 34.62±0.68 | 32.70±0.29 | 28.33±3.40 | 41.61±0.09 | 20.08±2.79 | <u>42.24±0.10</u> | **45.59±0.58** |
| | ARI | 34.99±0.14 | 33.58±1.19 | 33.12±0.52 | 25.62±5.05 | 42.58±0.09 | 14.45±2.38 | <u>43.03±0.17</u> | **46.89±0.49** |
| | F1 | 60.02±0.11 | 57.38±0.81 | 57.70±0.49 | 16.86±8.02 | 62.59±0.15 | 41.80±3.48 | <u>63.86±0.08</u> | **66.62±0.50** |
| DBLP | ACC | <u>66.85±0.35</u> | 61.20±1.21 | 58.59±0.08 | 53.83±9.17 | 65.96±0.18 | 46.09±3.84 | 64.27±0.31 | **78.18±0.83** |
| | NMI | <u>34.49±0.16</u> | 30.80±0.91 | 26.91±0.08 | 21.57±7.44 | 32.03±0.25 | 12.44±2.09 | 33.59±0.35 | **47.30±1.31** |
| | ARI | 29.19±0.91 | 22.01±1.40 | 17.91±0.09 | 17.54±10.47 | <u>34.76±0.23</u> | 10.30±2.46 | 28.83±0.27 | **51.94±1.62** |
| | F1 | 67.54±0.06 | 61.41±2.21 | 58.70±0.09 | 22.70±10.18 | 64.19±0.22 | 45.99±3.90 | <u>68.27±0.33</u> | **77.53±0.87** |
| HHAR | ACC | <u>75.68±0.31</u> | 62.31±1.00 | 71.30±0.38 | 62.33±6.68 | 54.24±0.27 | 60.48±4.01 | 73.46±0.32 | **83.07±0.09** |
| | NMI | <u>71.72±0.97</u> | 55.08±1.40 | 62.98±0.38 | 57.86±4.50 | 65.36±0.46 | 57.17±1.72 | 68.35±0.58 | **81.05±0.13** |
| | ARI | <u>64.28±0.88</u> | 42.62±1.61 | 51.48±0.71 | 47.14±5.73 | 47.32±0.25 | 44.98±2.77 | 60.00±0.40 | **73.45±0.01** |
| | F1 | <u>74.56±0.34</u> | 62.62±0.98 | 71.57±0.29 | 13.35±7.24 | 45.07±0.39 | 59.15±5.16 | 73.47±0.34 | **83.81±0.10** |
| Reuters | ACC | <u>79.35±0.21</u> | 54.40±0.29 | 60.88±0.22 | 59.32±6.31 | 60.27±0.30 | 49.66±5.08 | 50.35±7.36 | **81.53±0.67** |
| | NMI | <u>50.97±0.29</u> | 25.91±0.41 | 25.51±0.21 | 31.90±8.54 | 33.04±0.09 | 20.64±5.30 | 30.04±8.48 | **56.76±0.59** |
| | ARI | <u>59.23±0.37</u> | 19.61±0.21 | 26.19±0.38 | 30.68±9.20 | 30.31±0.59 | 18.24±5.06 | 17.19±9.09 | **64.77±0.96** |
| | F1 | <u>71.07±0.22</u> | 43.52±0.41 | 57.12±0.18 | 22.65±13.22 | 57.45±0.14 | 46.12±5.31 | 36.90±8.07 | **73.58±0.30** |

**Table 4.** Clustering results on seven datasets (mean ± std) – 2

| Dataset | Metric | TDCN | DFCN | SDCN | IDEC | SCGC | CONVERT | MBN | OURS |
|---------|--------|------|------|------|------|------|---------|-----|------|
| ACM | ACC | 90.82±0.19 | 90.70±0.31 | 87.26±0.95 | 85.11±0.51 | 89.84±0.44 | 86.35±1.54 | <u>93.01±0.13</u> | **93.39±0.29** |
| | NMI | 69.20±0.50 | 69.27±0.66 | 60.53±2.57 | 56.61±1.18 | 66.73±0.99 | 58.35±3.50 | <u>74.94±0.26</u> | **76.45±0.08** |
| | ARI | 74.84±0.46 | 74.62±0.75 | 66.20±2.33 | 62.18±1.50 | 72.48±1.03 | 63.89±3.71 | <u>80.39±0.33</u> | **81.44±0.07** |
| | F1 | 90.80±0.19 | 90.68±0.31 | 87.18±0.97 | 85.10±0.49 | 89.77±0.46 | 86.33±1.52 | <u>93.02±0.13</u> | **93.39±0.03** |

**Continued Table 4**

| | | | | | | | | | |
|---|---|---|---|---|---|---|---|---|---|
| **Cora** | ACC | - | 49.93±0.08 | 54.27±2.05 | 49.00±1.45 | 73.11±1.53 | **73.99±1.54** | 69.79±0.57 | <u>73.24±0.06</u> |
| | NMI | - | 26.59±0.16 | 34.40±0.78 | 28.83±1.70 | **55.78±0.80** | <u>55.50±1.11</u> | 48.81±1.03 | 55.16±0.14 |
| | ARI | - | 21.94±0.16 | 28.88±1.02 | 22.09±2.54 | <u>51.25±2.42</u> | 50.49±2.03 | 47.03±0.88 | **52.19±0.11** |
| | F1 | - | 48.38±0.08 | 37.34±3.54 | 48.85±1.77 | <u>70.01±2.31</u> | **72.84±3.26** | 64.89±0.63 | 64.79±0.04 |
| **CiteSeer** | ACC | 69.10±0.53 | 69.97±0.15 | 58.72±0.91 | 60.50±1.41 | 71.08±0.64 | 67.85±0.65 | <u>71.19±0.34</u> | **71.37±0.34** |
| | NMI | 41.61±0.65 | 44.21±0.19 | 32.27±0.59 | 27.19±2.40 | 45.20±0.36 | 41.02±0.85 | <u>45.56±0.67</u> | **45.59±0.58** |
| | ARI | 43.91±0.56 | 46.12±0.18 | 32.32±0.38 | 25.70±2.67 | 46.15±1.15 | 42.01±0.12 | **47.02±0.52** | <u>46.89±0.49</u> |
| | F1 | 62.30±0.31 | 64.69±0.17 | 52.29±4.92 | 61.61±1.40 | 61.92±0.57 | 62.17±1.88 | <u>65.67±0.27</u> | **66.62±0.50** |
| **DBLP** | ACC | - | 76.24±0.21 | 63.64±0.56 | 60.30±0.61 | 67.10±1.69 | 63.75±2.29 | <u>77.04±0.06</u> | **78.18±0.83** |
| | NMI | - | 44.02±0.24 | 34.44±0.60 | 31.19±0.50 | 38.32±1.80 | 27.92±2.51 | <u>46.95±0.11</u> | **47.30±1.31** |
| | ARI | - | 46.95±0.50 | 35.26±0.82 | 25.38±0.60 | 35.34±1.82 | 27.07±2.99 | <u>51.01±0.11</u> | **51.94±1.62** |
| | F1 | - | 76.04±0.19 | 62.75±0.47 | 61.31±0.58 | 66.95±1.80 | 63.78±2.32 | <u>76.34±0.06</u> | **77.53±0.87** |
| **HHAR** | ACC | **88.32±0.12** | <u>87.03±0.02</u> | 70.26±5.64 | 71.07±0.38 | 52.78±3.41 | 72.77±1.89 | 79.80±0.30 | 83.07±0.09 |
| | NMI | **82.24±0.40** | <u>82.17±0.04</u> | 71.47±5.53 | 74.19±0.40 | 52.34±3.84 | 67.87±1.88 | 77.11±0.30 | 81.05±0.13 |
| | ARI | **77.22±0.24** | <u>76.29±0.04</u> | 60.31±6.87 | 68.50±0.45 | 38.96±4.78 | 57.42±2.84 | 68.87±0.40 | 73.45±0.01 |
| | F1 | **88.12±0.15** | <u>87.30±0.03</u> | 66.86±7.29 | 68.61±0.32 | 49.43±4.68 | 72.49±2.08 | 80.07±0.34 | 83.81±0.10 |
| **Reuters** | ACC | <u>81.50±1.09</u> | 77.76±0.31 | 78.36±0.27 | 75.42±0.12 | 73.40±0.58 | 70.85±1.33 | 77.89±0.17 | **81.53±0.67** |
| | NMI | <u>59.28±0.97</u> | **59.82±0.80** | 55.93±0.27 | 50.29±0.18 | 50.26±2.57 | 43.08±2.10 | 50.58±0.38 | 56.76±0.59 |
| | ARI | <u>62.46±1.94</u> | 59.76±0.80 | 56.58±0.18 | 51.28±0.21 | 48.54±2.60 | 43.41±2.65 | 56.83±0.39 | **64.77±0.96** |
| | F1 | 66.28±0.23 | 69.57±0.25 | 63.42±0.80 | **76.21±0.11** | 61.24±1.75 | 66.96±1.47 | 70.48±0.17 | <u>73.58±0.30</u> |

Clustering performance (%) of our method and 14 comparison methods.

**Bold** values are the best results, <u>underlined</u> values are the second best results.

"-" indicates that the method is not mentioned.

Compared to traditional clustering models such as AE, our model demonstrates superior performance across four metrics on most datasets. GCL-GCN effectively captures graph structural information and node features through the GCN module, enabling a deeper understanding and representation of complex relationships in graph data, which traditional clustering methods typically fail to fully exploit. Consequently, GCL-GCN exhibits significant advantages when handling datasets with intricate relationships and highly nonlinear structures.

In comparison with other graph neural network-based clustering models (such as DFCN, DAEGC, ARGA, and MBN), GCL-GCN displays superior clustering performance. This success can be attributed to our model's AE module, GCN module, Graphormer module, and contrastive learning module. Notably, the Graphormer module allows GCL-GCN to capture long-range dependencies between nodes, enhancing its ability to grasp global information in complex graph structures. This results in better identification of inherent patterns and structures within the data, leading to improved information retention and more accurate clustering.

Compared to the Transformer-based clustering model TDCN, GCL-GCN enhances performance by combining the feature matrix processed through the Graphormer module, which employs centrality encoding, spatial encoding, and edge encoding, with the original feature matrix as input. Specifically, the Graphormer module extracts topological information from graph data using combined encoding methods and captures long-range dependencies through a multi-head attention mechanism. Additionally, our unique contrastive learning module further enhances feature discrimination, improving clustering outcomes. This multi-module integrated design enables GCL-GCN to exhibit higher stability and robustness when handling graph data, resulting in minimal result fluctuations. Compared to TDCN, GCL-GCN demonstrates superior clustering performance on most datasets and evaluation metrics due to its multi-module integration, enhanced feature discrimination, and effective capture of both global and local information. However, GCL-GCN's NMI metric on the HHAR dataset and various metrics on the Reuters dataset fall short of TDCN's performance. These lower metrics may be attributed to GCL-GCN's limited capability in handling highly dynamic and heterogeneous graph data, particularly in scenarios requiring the capture of subtle long-range dependencies. Although the Graphormer module effectively mitigates GCN's limitations in most cases, it may not fully realize its potential in these specific datasets.

Compared to contrastive learning-based clustering models such as CONVERT and SCGC, GCL-GCN significantly enhances the discriminative power of feature representations by employing contrastive learning techniques during the pre-training phase. The multi-task joint optimization strategy of GCL-GCN further improves the clustering performance of node representations, demonstrating greater stability and robustness across various types of datasets. This results in superior performance on most datasets compared to these models.

GCL-GCN shows remarkable performance improvements over traditional deep clustering models, graph neural network-based clustering models, Transformer-based deep clustering models, and contrastive learning-based clustering models. By effectively leveraging graph structural data, GCL-GCN achieves higher accuracy, stability, and robustness across multiple datasets, highlighting its potential for advanced clustering tasks in high-dimensional and complex data environments.

## 4.6 Hyperparametric analysis

In this section, we delve into the performance of different datasets under various weight configurations, focusing on the impact of hyperparameters $\lambda$, $\theta$, and $\gamma$ on the model's clustering performance. We evaluated these weight configurations using four performance metrics to assess their specific impact on performance. Through systematic parameter adjustments, we individually adjusted the weight parameters $\lambda$, $\theta$, and $\gamma$ for the GCN, AE, and Graphormer models. We set $\lambda + \theta + \gamma = 1$ and conducted a quantitative analysis of the hyperparameters' effects using performance metrics obtained from six datasets. To clearly illustrate the impact of parameter variations on performance, we generated heatmaps for each dataset's $\lambda$ and $\theta$ parameters (through $\lambda$ and $\theta$, we can determine $\gamma$), as shown in Figure 6, which visually depict the effect of these parameters on the F1 scores across different datasets. Based on these visualizations, we summarized the highest metrics achieved for each dataset under specific hyperparameter configurations. The results indicate that the hyperparameters $\lambda$, $\theta$, and $\gamma$ have varying impacts on performance metrics across different datasets, further demonstrating that different datasets have distinct requirements for hyperparameter configurations.

**Impact of Parameter Variations on Clustering Performance**

As shown in Table 5, performance metrics for different datasets achieved optimal levels under specific parameter settings. The results indicate that the model can attain the best performance with tailored parameter configurations based on the inherent attributes of different datasets.

The dependency of different datasets on $\lambda$, $\theta$, and $\gamma$ varies due to differences in their feature complexity, graph structure, and task requirements. For example, the ACM dataset, which primarily contains academic papers and their citation relationships, exhibits a balanced weight distribution for $\lambda$, $\theta$, and $\gamma$ due to its rich features, local structural information, and global relationships. In contrast, the Cora and Reuters datasets, with their complex relationships, rely more heavily on the Graphormer module ($\gamma$ is higher) because the intricate citation or semantic relationships between documents or news texts require capture through a global attention mechanism.

The DBLP dataset, characterized by complex features, shows a higher dependency on the GCN module ($\lambda$ is higher), indicating the significant role of GCN in feature extraction and compression when dealing with complex academic publications and their author relationships. For tasks where neighborhood information is crucial, such as the CiteSeer dataset, there is a higher dependency on the AE module ($\theta$ is higher), demonstrating that AE's propagation of local neighborhood information is vital for citation networks. Similarly, the HHAR dataset also relies more on the AE module ($\theta$ is higher), underscoring AE's effectiveness in handling such data.

**Table 5.** Highest F1 scores and their corresponding parameter sets in the 6 datasets (in this experiment, we set both $\boldsymbol{\alpha}$ and $\boldsymbol{\beta}$ to 0.1)

| Dataset | $\lambda$ | $\theta$ | $\gamma$ | The highest F1 score |
|---------|-----------|----------|----------|----------------------|
| ACM | 0.4 | 0.3 | 0.3 | 0.9382 |
| Cora | 0.4 | 0.1 | 0.5 | 0.6489 |
| CiteSeer | 0.1 | 0.8 | 0.1 | 0.6718 |

**Continued Table 5**

| | | | | |
|---|---|---|---|---|
| DBLP | 0.7 | 0.1 | 0.2 | 0.7922 |
| HHAR | 0.1 | 0.8 | 0.1 | 0.8403 |
| Reuters | 0.4 | 0.1 | 0.5 | 0.7380 |

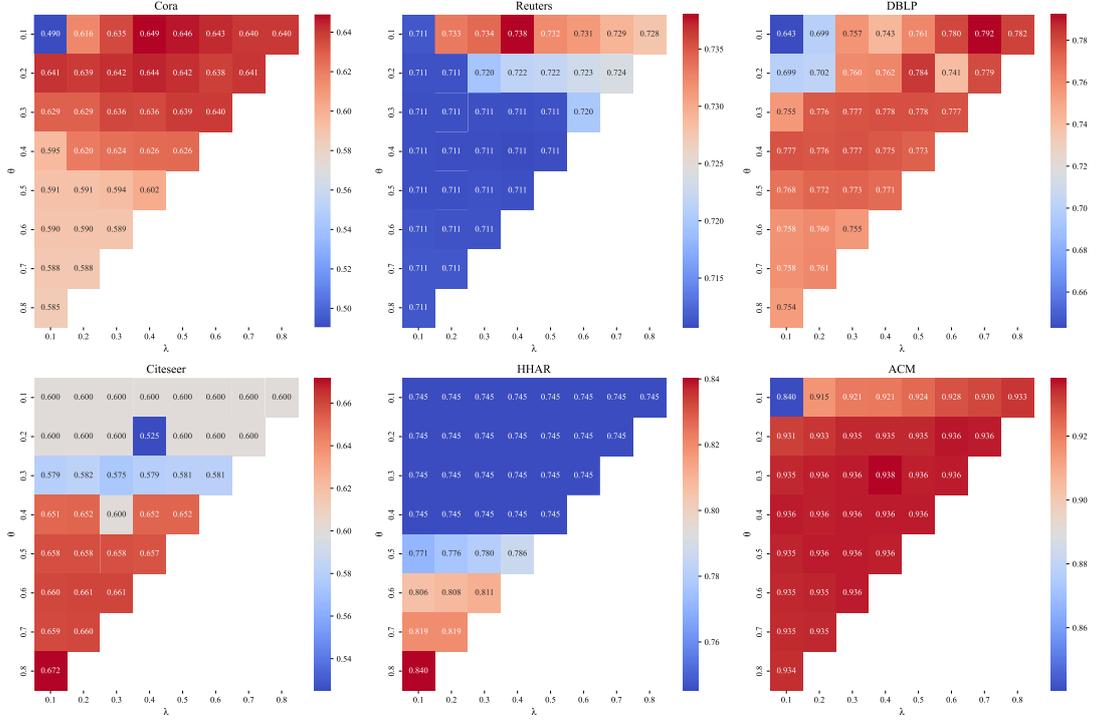

**Fig. 6.** Heatmap of six datasets showing the effect of $\lambda$ and $\theta$ parameters on the impact of F1 scores on different datasets

### 4.6.1 Hyperparameter Analysis of Total Loss Function Weights

In this experiment, we analyzed the impact of hyperparameters $\boldsymbol{\alpha}$ and $\boldsymbol{\beta}$ on model performance. Here, $\boldsymbol{\alpha}$ represents the weight of the self-supervised clustering loss, and $\boldsymbol{\beta}$ represents the weight of the consistency constraint loss. The values of $\boldsymbol{\alpha}$ and $\boldsymbol{\beta}$ are taken from {0.01, 0.05, 0.08, 0.1, 0.12, 0.15, 0.3}. We evaluated their effects on the model's performance across six datasets and visualized the results in Figure 7.

The experimental results indicate that different combinations of hyperparameters significantly affect the model's performance on various datasets. For instance, on the ACM and Reuters datasets, larger values of $\boldsymbol{\alpha}$ and $\boldsymbol{\beta}$ better capture the intrinsic characteristics of the data, thereby enhancing model performance. Conversely, on the Cora dataset, smaller values of $\boldsymbol{\alpha}$ and $\boldsymbol{\beta}$ better balance feature extraction, yielding the best results. By selecting appropriate values for $\boldsymbol{\alpha}$ and $\boldsymbol{\beta}$ based on the specific characteristics of different datasets, we can significantly improve the model's performance.

As shown in Figure 7, the impact of varying $\alpha$ and $\beta$ on model performance across different datasets is illustrated. We found that for most datasets, large variations in $\alpha$ and $\beta$ resulted in minimal changes in model performance. The bar charts on the 3D plot exhibit relatively flat changes for most datasets, suggesting that our model is not particularly sensitive to the $\alpha$ and $\beta$ parameters.

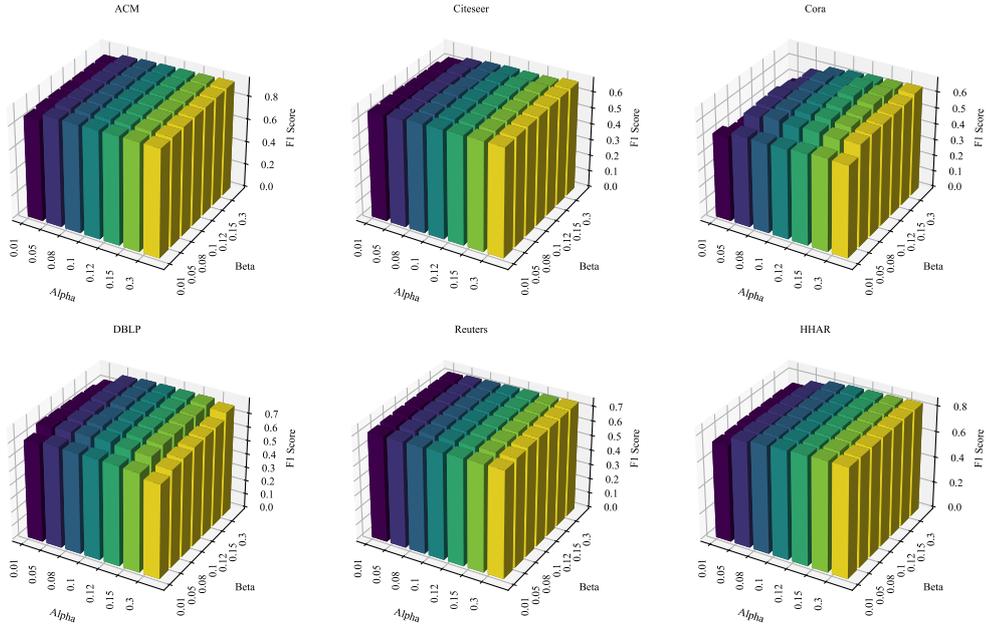

**Fig. 7.** 3D plot showing the effect of $\alpha$ and $\beta$ parameters on the impact of F1 scores on different datasets

## 4.7 Ablation Study

By comparing the original model ("norm") with versions where each of the three modules was separately removed, we conducted ablation experiments on six datasets. We then compared the performance across four metrics to evaluate the contribution of each module within the model. Detailed data can be found in Table 6.

We separately removed the GCN module, Graphormer module, and Contrastive Learning module from the complete model. The complete model achieved the highest scores across most metrics, indicating the best overall performance. Upon removing the GCN module, there was a significant drop in performance, with the average ACC decreasing to 0.7389, NMI to 0.5195, ARI to 0.5229, and F1 to 0.7165. The Graphormer and Contrastive Learning modules also proved to be crucial; their removal led to the loss of the self-attention mechanism and a decrease in feature discrimination capability, respectively, resulting in performance declines across all four metrics. These findings underscore the importance of the GCN, Graphormer, and Contrastive Learning modules in our model.

**Table 6.** Results of module ablation analysis on six datasets based on four metrics.
norm: indicates the results obtained without modifying the model.

| Metrics | ACC | | | | NMI | | | |
|---|---|---|---|---|---|---|---|---|
| Operation Dataset | norm | -GCN | -Graphormer | -Contrastive Learning | norm | -GCN | -Graphormer | -Contrastive Learning |
| ACM | **0.9382** | 0.9342 | **0.9382** | <u>0.9375</u> | <u>0.7750</u> | 0.7634 | <u>0.7750</u> | **0.7756** |
| CiteSeer | **0.7181** | <u>0.7178</u> | <u>0.7178</u> | **0.7181** | **0.4639** | <u>0.4632</u> | 0.4616 | **0.4639** |
| Cora | **0.7352** | 0.4793 | <u>0.4793</u> | <u>0.4793</u> | **0.5572** | <u>0.2711</u> | <u>0.2711</u> | <u>0.2711</u> |
| DBLP | **0.7981** | 0.6734 | 0.7873 | <u>0.7962</u> | **0.4969** | 0.2983 | 0.4735 | <u>0.4914</u> |
| HHAR | <u>0.8327</u> | **0.8343** | 0.8320 | 0.8321 | <u>0.8125</u> | **0.8131** | 0.8112 | 0.8119 |
| Reuters | **0.8326** | 0.7944 | <u>0.8261</u> | 0.8239 | **0.5789** | 0.5079 | <u>0.5723</u> | 0.5636 |
| Metrics | ARI | | | | F1 | | | |
| Operation Dataset | norm | -GCN | -Graphormer | -Contrastive Learning | norm | -GCN | -Graphormer | -Contrastive Learning |
| ACM | **0.8254** | 0.8149 | **0.8254** | <u>0.8239</u> | **0.9382** | 0.9342 | **0.9382** | <u>0.9375</u> |
| CiteSeer | **0.4748** | <u>0.4743</u> | 0.4739 | **0.4748** | **0.6719** | 0.6712 | 0.6713 | <u>0.6718</u> |
| Cora | **0.5307** | <u>0.1975</u> | <u>0.1975</u> | <u>0.1975</u> | **0.6489** | <u>0.4770</u> | <u>0.4770</u> | <u>0.4770</u> |
| DBLP | **0.5495** | 0.3202 | 0.5246 | <u>0.5443</u> | **0.7922** | 0.6641 | 0.7813 | <u>0.7909</u> |
| HHAR | <u>0.7372</u> | **0.7381** | 0.7362 | 0.7364 | <u>0.8403</u> | **0.8417** | 0.8394 | 0.8397 |
| Reuters | **0.6672** | 0.5924 | <u>0.6574</u> | 0.6506 | <u>0.7380</u> | 0.7109 | 0.7374 | **0.7415** |

-GCN: indicates the result after omitting the GCN module.

-Graphormer: indicates the result after omitting Graphormer.

-Contrastive Learning: indicates the results after omitting Contrastive Learning.

**Bold** values represent the highest data and <u>underlined</u> values represent the second highest data.

## 4.8 Layer Number Experiment

To verify the impact of different propagation layers on our model, we configured the model with 1, 2, 3, and 4 layers and evaluated its performance on six datasets using four metrics. As shown in Table 7, GCL-GCN-3 achieved the best results across all four metrics on all six datasets. This indicates that a three-layer propagation structure is more effective for feature extraction, enhancing the model's clustering performance. Therefore, we selected three encoding layers and three decoding layers for the network components of each module. The results for GCL-GCN-1 and GCL-GCN-2 demonstrate that having fewer network layers limits the model's learning capacity, preventing effective feature extraction. On the other hand, the performance of GCL-GCN-4 significantly declined, likely due to overfitting as the number of network layers increased, leading to a reduction in clustering performance.

**Table 7.** Results of different propagation layer numbers on 6 datasets across 4 metrics

| Dataset | Propagation | ACC | NMI | ARI | F1 |
|---------|-------------|-----|-----|-----|-----|
| **ACM** | GCL-GCN-4 | 73.59 | 41.47 | 43.72 | 72.03 |
| | GCL-GCN-3 | **93.82** | **77.50** | **82.54** | **93.82** |
| | GCL-GCN-2 | 93.29 | 75.70 | 81.13 | 93.28 |
| | GCL-GCN-1 | 91.21 | 71.01 | 75.87 | 91.20 |
| **DBLP** | GCL-GCN-4 | 44.49 | 19.64 | 9.89 | 40.18 |
| | GCL-GCN-3 | **79.81** | **49.69** | **54.95** | **79.22** |
| | GCL-GCN-2 | 44.59 | 23.31 | 14.41 | 39.26 |
| | GCL-GCN-1 | 76.53 | 47.22 | 47.26 | 76.31 |
| **CiteSeer** | GCL-GCN-4 | 51.16 | 27.83 | 26.19 | 47.91 |
| | GCL-GCN-3 | **71.81** | **46.39** | **47.48** | **67.18** |
| | GCL-GCN-2 | 69.07 | 43.84 | 45.42 | 64.62 |
| | GCL-GCN-1 | 58.43 | 30.93 | 30.76 | 54.55 |
| **HHAR** | GCL-GCN-4 | 80.63 | 70.71 | 64.11 | 80.20 |
| | GCL-GCN-3 | **83.27** | **81.25** | **73.72** | **84.03** |
| | GCL-GCN-2 | 76.35 | 79.08 | 68.32 | 73.61 |
| | GCL-GCN-1 | 71.26 | 79.83 | 67.41 | 69.94 |
| **Reuters** | GCL-GCN-4 | 70.35 | 44.92 | 49.65 | 63.77 |
| | GCL-GCN-3 | **83.26** | **57.89** | **66.72** | **73.80** |
| | GCL-GCN-2 | 73.59 | 56.55 | 59.13 | 66.82 |
| | GCL-GCN-1 | 57.35 | 42.59 | 35.15 | 48.73 |
| **Cora** | GCL-GCN-4 | 44.57 | 30.26 | 10.30 | 32.86 |
| | GCL-GCN-3 | **73.52** | **55.72** | **53.07** | **64.89** |
| | GCL-GCN-2 | 72.82 | 55.44 | 52.19 | 64.00 |
| | GCL-GCN-1 | 60.49 | 45.72 | 39.54 | 53.50 |

Percentages are used to express the clustering results of our model on the four versions.

GCL-GCN-x means there are x layers of encoders or decoders.

**Bold** values are the best results.

## 4.9 Analysis of Different Encoding Methods

To verify the effectiveness of our encoding design, we proposed several experimental methods. Firstly, in spatial encoding, we replaced the spatial measure from Euclidean Distance (ED) to Shortest Path Distance (SPD) to validate the effectiveness of the spatial encoding design. Secondly, in centrality encoding, we evaluated the effectiveness by replacing the combined use of Degree Centrality (DC), Closeness Centrality (CC), and Betweenness Centrality (BC) with the use of only DC, CC, or BC individually. As shown in Table 8, based on tests across six datasets and four metrics, demonstrate that our proposed encoding design performs better compared to simply using SPD or a single centrality measure. Overall, our model's encoding design exhibits superior performance across the majority of datasets and evaluation metrics, validating its effectiveness and applicability.

**Table 8.** Results of different encoding methods on 6 datasets across 4 metrics

| Metric \ Dataset | | ACM | DBLP | CiteSeer | Cora | HHAR | Reuters |
|---|---|---|---|---|---|---|---|
| **GCL-GCN** | ACC | **0.9382** | **0.7981** | **0.7184** | **0.7345** | 0.8322 | **0.8326** |
| | NMI | **0.7754** | **0.4969** | 0.4636 | **0.5549** | **0.8138** | **0.5789** |
| | ARI | **0.8254** | **0.5501** | **0.4751** | **0.5292** | 0.7381 | **0.6672** |
| | F1 | **0.9382** | **0.7912** | **0.6720** | **0.6480** | 0.8399 | 0.7380 |
| **DC, BC and CC + SPD** | ACC | 0.9372 | 0.7954 | 0.7181 | 0.4793 | 0.8325 | 0.8239 |
| | NMI | 0.7742 | 0.4912 | **0.4639** | 0.2711 | 0.8123 | 0.5638 |
| | ARI | 0.8230 | 0.5424 | 0.4748 | 0.1975 | 0.7369 | 0.6508 |
| | F1 | 0.9372 | 0.7903 | 0.6718 | 0.4770 | **0.8401** | **0.7415** |
| **DC + ED** | ACC | 0.9372 | 0.7966 | 0.7172 | 0.5905 | 0.8322 | 0.8253 |
| | NMI | 0.7717 | 0.4912 | 0.4622 | 0.4697 | 0.8132 | 0.5739 |
| | ARI | 0.8227 | 0.5453 | 0.4731 | 0.3866 | 0.7376 | 0.6589 |
| | F1 | 0.9372 | 0.7910 | 0.6706 | 0.5300 | 0.8399 | 0.7397 |
| **BC + ED** | ACC | 0.9372 | 0.7934 | 0.7172 | 0.4793 | 0.8337 | 0.8253 |
| | NMI | 0.7717 | 0.4883 | 0.4622 | 0.2711 | 0.8137 | 0.5741 |
| | ARI | 0.8227 | 0.5430 | 0.4731 | 0.1975 | **0.7385** | 0.6589 |
| | F1 | 0.9372 | 0.7862 | 0.6706 | 0.4770 | 0.8412 | 0.7397 |
| **CC + ED** | ACC | 0.9372 | 0.7934 | 0.7172 | 0.4793 | **0.8338** | 0.8253 |
| | NMI | 0.7717 | 0.4883 | 0.4622 | 0.2711 | 0.8132 | 0.5741 |
| | ARI | 0.8227 | 0.5430 | 0.4731 | 0.1975 | 0.7380 | 0.6589 |
| | F1 | 0.9372 | 0.7862 | 0.6706 | 0.4770 | 0.8413 | 0.7397 |

**Bold** value are the highest values for the same metric from different experiments in the same dataset.

**GCL-GCN:** DC, BC and CC for central coding measures and ED for spatial coding measures.

**DC, BC and CC + SPD:** DC, BC and CC are used for centrally coded measures and SPD for spatially coded measures.

## 5. Discussion

**Discussion 1: Sensitivity of Hyperparameters**

Hyperparameters directly influence the model's performance, training efficiency, and generalization ability. Proper hyperparameter settings can enhance the model's accuracy and robustness, expedite training, and effectively control the model's complexity to prevent overfitting [50]. Therefore, we conducted experiments to examine the sensitivity of hyperparameters $\lambda$ and $\theta$, calculating the composite index across six datasets ( Composite index $= \frac{ACC+NMI+ARI+F1}{4}$ ). The results are illustrated in Figure 8. In the figure, the width of the green shaded areas indicates sensitivity to variations in $\lambda$ and $\theta$. Wider shaded areas suggest higher sensitivity, while narrower areas indicate lower sensitivity, demonstrating greater stability. The sensitivity to changes in λ and θ varies across different datasets. From the figure, it is evident that the HHAR dataset shows low sensitivity to changes in θ. The composite index remains relatively stable until θ = 0.4, after which it starts to rise gradually. Conversely, Reuters and ACM are more affected by θ changes before 0.4, with their composite indices leveling off afterward. Compared to θ, the model is more sensitive to changes in λ. The HHAR dataset shows a significant decline in its composite index with increasing λ until it stabilizes after 0.5. Although Cora and ACM exhibit lower stability before 0.6 and 0.5, respectively, their composite indices generally trend upward, showing reduced sensitivity and gradual stabilization afterward. While this study provides insights into the sensitivity of hyperparameters λ and θ across different datasets, further research is needed to delve into automated hyperparameter optimization. In future work, we plan to develop and validate automated hyperparameter optimization techniques to ensure the model's stability and superior performance across various complex data environments.

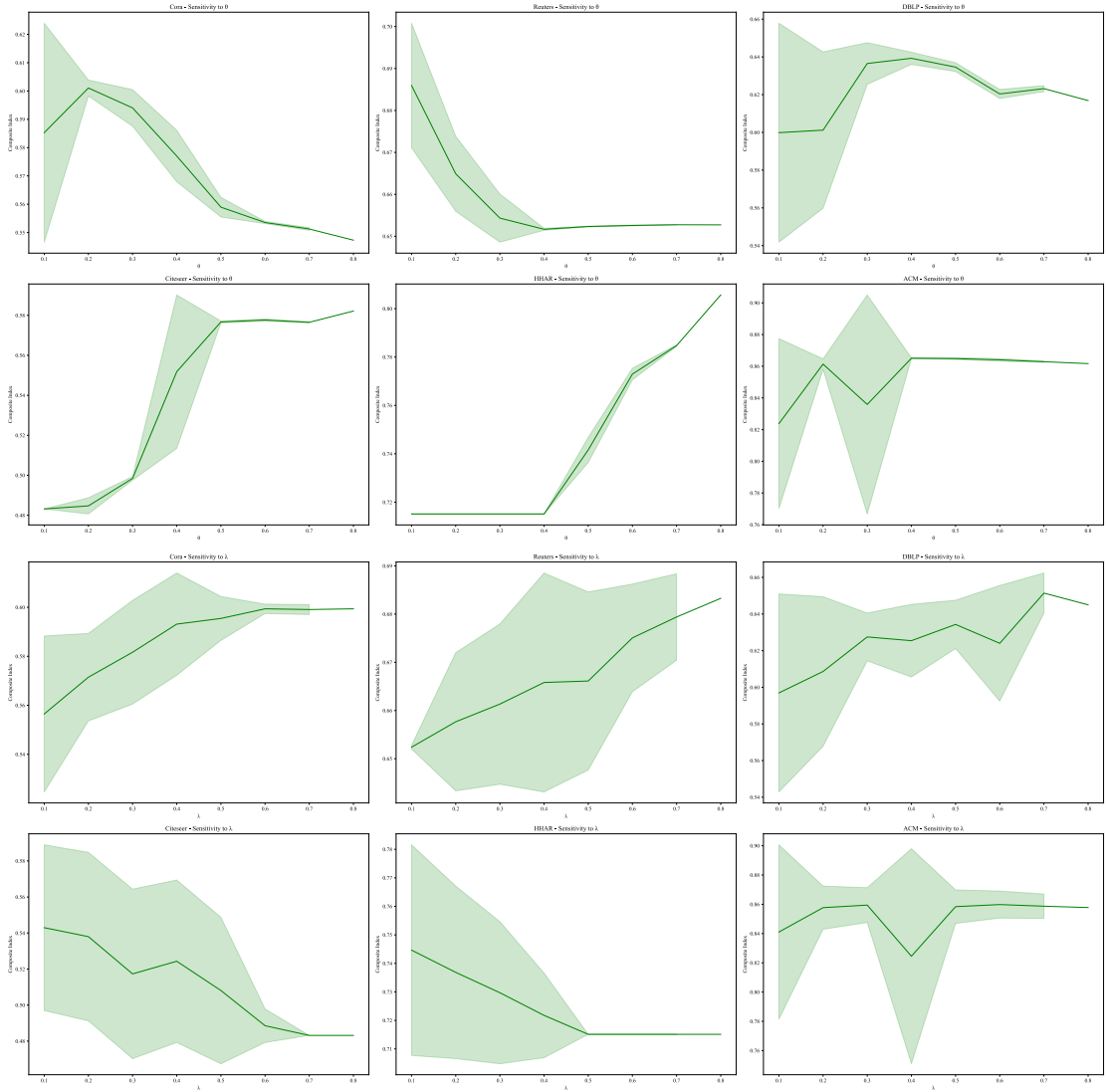

**Fig. 8.** The above figure shows the results of the hyperparameter sensitivity analysis on six datasets,

the former for the model's sensitivity to θ and the latter for the sensitivity to λ.

**Discussion 2: Is a More Complex Contrastive Learning Method More Effective?**

In our related work, we mentioned that using a contrastive learning module with a complex architecture might lead to a decline in clustering performance. To validate this hypothesis, we designed an experiment to compare the performance of different contrastive learning module designs when applied to our model. In this experiment, we compared the performance of three different contrastive models: v0 (the model used in this paper), v1, and v2. Specifically, the v0 model employs a two-layer GCN combined with a simple random dropout data augmentation method. In contrast, the v1 model builds on v0 by introducing more complex data augmentation strategies, including a combination of Gaussian noise, feature cropping, and normalization. The v2 model, on the other hand, increases the complexity by using a four-layer GCN based on the v0 architecture. Due to the unstable performance of v1 on the Cora and HHAR datasets, as well as v2 on the HHAR dataset, we chose to analyze the highest scores of the evaluation metrics on these datasets. As shown in Figure 9, v0 performs excellently on the ACM, Cora, and HHAR datasets, possibly because its shallow architecture better balances model complexity and data noise. In contrast, v1, which uses complex data augmentation strategies, showed a decline in performance on some datasets such as Cora and HHAR. This decline might be due to the over-augmentation disturbing the features. Although v2 showed improvements on some datasets, its overall performance was not as good as v0, particularly on the Cora dataset, where the complex model amplified data noise, resulting in inaccurate feature extraction. Through our experiments, we found that the shallow architecture of v0 performs best in most cases.

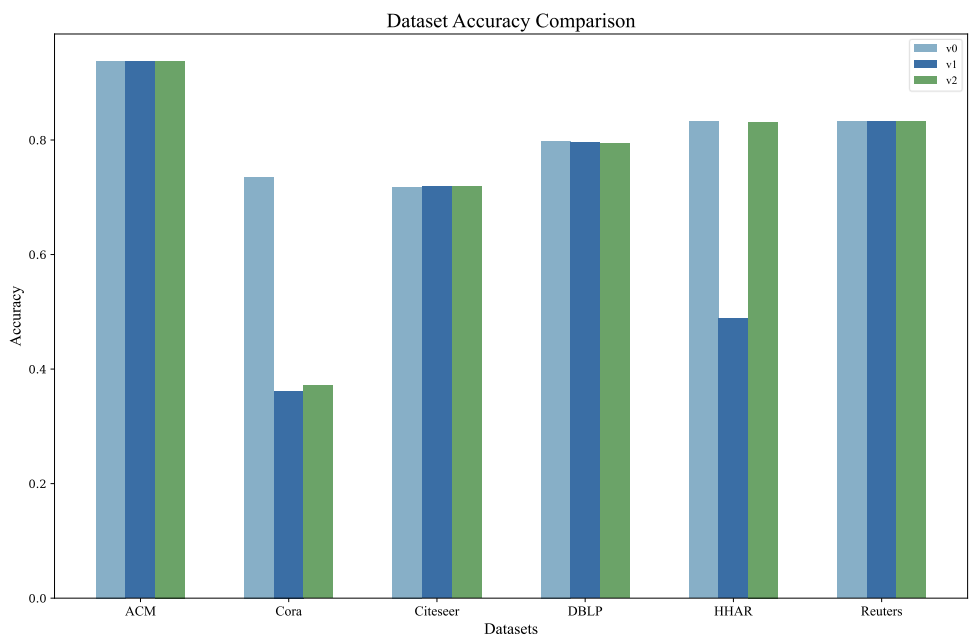

**Fig. 9.** Comparison of model performance using different comparative learning module versions.

## 6. Conclusion

In this study, we proposed GCL-GCN, an innovative deep graph clustering model featuring a newly designed Graphormer with novel encoding methods and a new contrastive learning module. GCL-GCN leverages a modular design to effectively utilize the advantages of the AE, GCN, Graphormer, and contrastive learning modules, achieving efficient feature extraction and clustering for graph data. Notably, the Graphormer module significantly enhances node representation and clustering performance by introducing centrality features and spatial relationship encoding. Additionally, contrastive learning pre-training markedly improves the discriminative power of feature representations. Extensive experiments on six benchmark datasets demonstrate that our model surpasses 14 state-of-the-art methods in overall clustering accuracy and stability. These results highlight GCL-GCN's significant advantages in handling complex graph data and capturing both global and local information, validating its potential in both theoretical and practical applications. Future research will focus on further exploring the application of GCL-GCN to other types of graph data and optimizing the model's performance, especially when dealing with larger-scale and more complex structures. GCL-GCN provides an effective and powerful tool for clustering analysis of attributed graph data, with broad application prospects.

## Author Contributions

**Binxiong Li:** Conceptualization, Methodology, Software, Validation, Formal analysis, Writing-Original draft, Review & Editing. **Xue Li:** Methodology, Software, Validation, Investigation, Data curation, Writing-Original draft, Review & Editing. **Xu Xiang:** Methodology, Software, Validation, Investigation, Data curation, Writing-Original draft, Review & Editing. **Quanzhou Lou:** Methodology, Software, Writing-Original draft**. Binyu Zhao:** Methodology, Software, Writing-Original draft. **Heyang Gao:** Methodology, Software, Writing-Original draft. **Yujie Liu:** Methodology, Software, Data curation. **Huijie Tang:** Validation, Software, Data curation. **Benhan Yang:** Validation, Data curation.

## Acknowledgments

The authors thank the editors and the anonymous referees for their valuable comments and efforts.

## Declarations

**Ethical Approval** Not applicable

**Competing interests** The authors declare no competing interests